\documentclass[sigconf, nonacm]{acmart}
\usepackage{multirow}
\usepackage{amsmath}
\usepackage[gen]{eurosym}
\usepackage{hyperref}
\usepackage{enumitem}
\setlist[itemize]{noitemsep, topsep=0pt}
\usepackage{graphicx}
\usepackage{caption}
\usepackage{comment}
\usepackage{subcaption}
\usepackage[utf8]{inputenc}
\usepackage[linesnumbered,ruled,vlined,algo2e]{algorithm2e}
\usepackage{multirow}
\usepackage{subcaption}
\AtBeginDocument{%
  \providecommand\BibTeX{{%
    \normalfont B\kern-0.5em{\scshape i\kern-0.25em b}\kern-0.8em\TeX}}}
\usepackage{balance }

\begin{document}
\def\x{{\mathbf x}}
\def\L{{\cal L}}
\def\eg{\textit{e.g.}}
\def\ie{\textit{i.e.}}
\def\Eg{\textit{E.g.}}
\def\etal{\textit{et al.}}
\def\etc{\textit{etc}}
\newcommand{\blue}[1]{\textcolor{blue}{#1}}

\title{Explainable  Depression Detection via Head Motion Patterns}

%

\author{Monika Gahalawat}
\affiliation{University of Canberra}
\email{monika.gahalawat@canberra.edu.au}

\author{Raul Fernandez Rojas}
\affiliation{ University of Canberra}
\email{raul.fernandezrojas@canberra.edu.au}

\author{Tanaya Guha}
\affiliation{University of Glasgow}
\email{tanaya.guha@glasgow.ac.uk}

\author{Ramanathan Subramanian}
\affiliation{University of Canberra}
\email{ram.subramanian@canberra.edu.au}

\author{Roland Goecke}
\affiliation{University of Canberra}
\email{roland.goecke@canberra.edu.au}

%


\begin{abstract}
While depression has been studied via multimodal non-verbal behavioural cues, head motion behaviour has not received much attention as a biomarker. This study demonstrates the utility of fundamental head-motion units, termed \emph{kinemes}, for depression detection by adopting two distinct approaches, and employing distinctive features: (a) discovering kinemes from head motion data corresponding to both depressed patients and healthy controls, and (b) learning kineme patterns only from healthy controls, and computing statistics derived from reconstruction errors for both the patient and control classes. Employing machine learning methods, we evaluate depression classification performance on the \emph{BlackDog} and \emph{AVEC2013} datasets. Our findings indicate that: (1) head motion patterns are effective biomarkers for detecting depressive symptoms, and (2) explanatory kineme patterns consistent with prior findings can be observed for the two classes. Overall, we achieve peak F1 scores of 0.79 and 0.82, respectively, over BlackDog and AVEC2013 for binary classification over episodic \emph{thin-slices}, and a peak F1 of 0.72 over videos for AVEC2013.   
\end{abstract}
\vspace{-3mm}

\vspace{-2mm}
\keywords{Kinemes, Head-motion, Depression detection, Explainability}


\maketitle
\vspace{-3.5mm}
\section{Introduction}\label{Sec:Intro}
Clinical depression, a prevalent mental health condition, is considered as one of the leading contributors to the global health-related burden \cite{greenberg2015economic, lepine2011increasing}, affecting millions of people worldwide \cite{vos_et_al_GBD2016,institute2021global}. As a mood disorder, it is characterised by a prolonged (> two weeks) feeling of sadness, worthlessness and hopelessness, a reduced interest and a loss of pleasure in normal daily life activities, sleep disturbances, tiredness and lack of energy. Depression can lead to suicide in extreme cases \cite{goldney2000suicidal} and is often linked to comorbidities such as anxiety disorders, substance abuse disorders, hypertensive diseases, metabolic diseases, and diabetes \cite{steffen_et_al_2020_BMCPsychiatry,campayo2011diabetes}. Although effective treatment options are available, diagnosing depression through self-report and clinical observations presents significant challenges due to the inherent subjectivity and biases involved.

Over the last decade, researchers from affective computing and psychology have focused on investigating objective measures that can aid clinicians in the initial diagnosis and monitoring of treatment progress of clinical depression \cite{cohn2018multimodal, pampouchidou_et_al_TAC_DepressionReview}. A key catalyst to this progress is the availability of relevant datasets, such as AVEC2013 and subsequent challenges~\cite{valstar2013avec}. In recent years, research on depression detection employing affective computing approaches has increasingly focused on leveraging non-verbal behavioural cues such as facial expressions \cite{bourke2010processing, de2019combining}, body gestures \cite{joshi2013relative}, eye gaze \cite{alghowinem2016multimodal}, head movements \cite{alghowinem2013head} and verbal features \cite{cummins2011investigation, huang2019investigation} extracted from multimedia data to develop distinctive features to classify individuals as depressed or healthy controls, or to estimate the severity of depression on a continuous scale. 

\begin{figure*}[t]
      \centering
     \includegraphics[width=\linewidth]{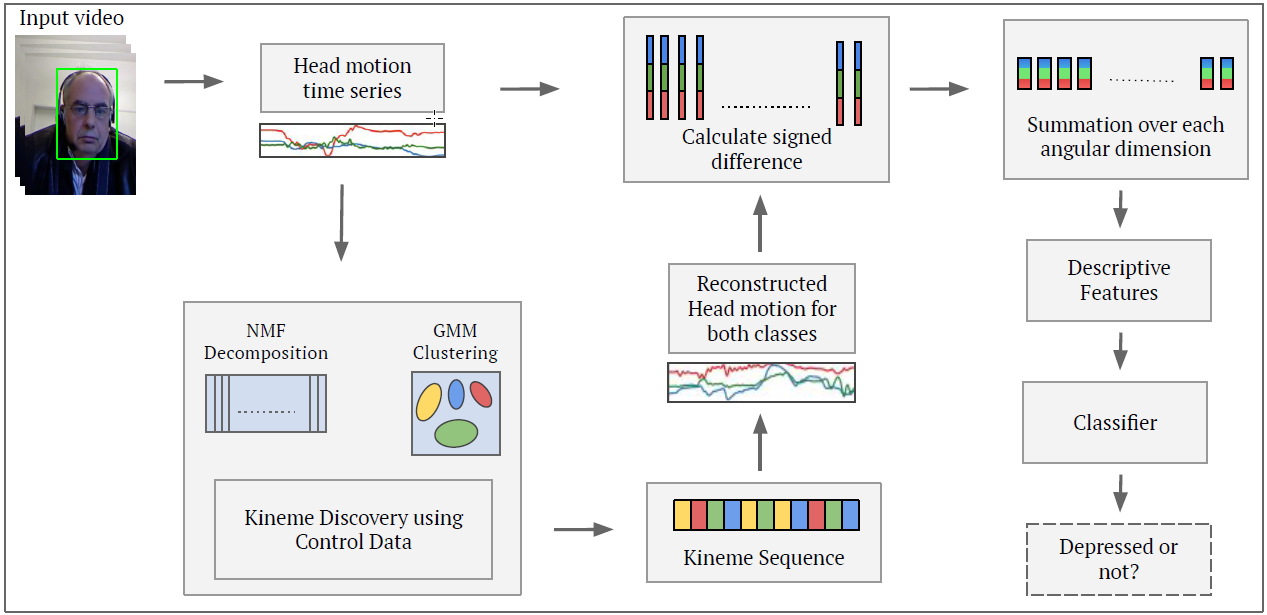} \vspace{-5mm}
    \caption{\textbf{Overview:}  \textmd{We learn kinemes for the control class, and the reconstruction errors between the raw and reconstructed head-motion segments, obtained via kineme clustering, are computed for both the control and depressed classes. Statistical descriptors over the yaw, pitch and roll dimensions (a total of $8 \times 3$ features) are utilized for depression detection via machine learning techniques. } } \label{fig:Depression_proposed_framework}\vspace{-4mm}
\end{figure*}

In this study, we examine the utility of inherently interpretable head motion units, referred to as \emph{kinemes} \cite{madan_gahalawat_guha_subramanian_ICMI2021_Kinemes}, for assessing depression. Initially, we utilise data from both healthy controls and depressed patients to discover a basis set of kinemes via the (\emph{pitch}, \emph{yaw}, and \emph{roll}) head pose angular data obtained from short overlapping time-segments (termed two-class kineme discovery or 2CKD). Further, we employ these kinemes to generate features based on the frequency of occurrence of distinctive, class-characteristic kinemes. Subsequently, we discover kineme patterns solely from head pose data corresponding to healthy controls (Healthy control kineme discovery or HCKD), and use them to represent both healthy and depressed class segments. A set of statistical features are then computed from the reconstruction errors between the raw and learned head-motion segments corresponding to both the depressed and control classes (see Figure ~\ref{fig:Depression_proposed_framework}). Using machine learning methodologies, we evaluate the performance of the features derived from the two approaches. Our results show that head motion patterns are effective behavioural cues for detecting depression. Additionally, explanatory class-specific kinemes patterns can be observed, in alignment with prior research.  

This paper makes the following research contributions:
\begin{itemize}
    \item A study of head movements as a biomarker for clinical depression, which so far has been understudied.
    \item Proposing the \textit{kineme} representation of motion patterns as an effective and explanatory means for depression analysis.
    \item \begin{sloppypar} A detailed investigation of various classifiers for 2-class and 4-class categorisation on the AVEC2013 and BlackDog datasets. We obtain peak F1-scores of 0.79 and 0.82, respectively, on \textit{thin-slice} chunks for binary classification on the BlackDog and AVEC2013 datasets, which compare favorably to prior approaches. Also, a video-level F1-score of 0.72 is achieved for 4-class categorisation on AVEC2013.  \end{sloppypar}
\end{itemize}
The remainder of this paper is organised as follows. Section \ref{Sec:RW} provides an overview of related work. Section \ref{Sec:KF} describes the kineme formulation, followed by Section \ref{Sec:EKF} that details the explainable kineme features used as a representation of motion patterns. The methodology is presented in Section \ref{Sec:Meth}, while Section \ref{Sec:ER} provides details of the datasets, experimental settings, and classifiers used in this study. The experimental results are shown and discussed in Section \ref{sec:ResultsDiscussion}. Finally, the conclusions are drawn in Section \ref{Sec:DC}.


\section{Related Work}\label{Sec:RW}

In this section, we briefly review the literature focusing on (a) depression detection as a classification problem, and (b) depression detection using head motion patterns.
%
\subsection{Depression Analysis as a Classification Task}
Traditionally, depression detection has been approached as a supervised binary classification task, with many studies relying on discriminative classifiers to distinguish between \emph{healthy controls} and \emph{patients} \cite{alghowinem2015cross, cohn2018multimodal, alghowinem2013head}. A typical recognition accuracy of up to 80\% demonstrates the promise of behavioural cues such as eye-blink and closed-eye duration rate, statistical features computed over the yaw, pitch and roll head-pose features, \etc. to differentiate the two classes. However, challenges involved in depression detection such as limited clinically validated, curated data and a skewed data distributions have been acknowledged in the literature~\cite{alghowinem2013head, nasir2016multimodal}. 

Recent efforts have sought to learn patterns indicative of only the target class and reformulate depression detection as a one-class classification problem to mitigate the issues with imbalanced datasets~\cite{opoku2019towards, aguilera2021depression}. Studies have attempted to learn features associated with control participants and treat inputs that deviate from these patterns as \textit{anomalous}~\cite{gerych2019classifying, mourao2011patient}. Gerych \emph{et. al.}~\cite{gerych2019classifying} formulate the task as anomaly detection by leveraging autoencoders to learn features of the non-depressed class and treating depressed user data as outliers. Similarly, Mourão-Miranda \emph{et. al.}~\cite{mourao2011patient} employ a one-class SVM to classify patients as outliers compared to healthy participants based on the fMRI responses to sad facial expressions. Conversely, a few studies explore one-class classification by learning features characterising the depressed class, and treating non-depressed subjects as outliers~\cite{aguilera2021depression, opoku2019towards}.

%
%
\subsection{Depression Detection via Head Motion Cues}
Many studies have focused on non-verbal behavioural cues, such as body gestures~\cite{joshi2013can, joshi2013relative}, facial expressions~\cite{bourke2010processing, he2022intelligent, de2019combining}, their combination~\cite{Parekh2018} and speech features~\cite{cummins2011investigation, rejaibi2022mfcc, huang2019investigation} as biomarkers for depression diagnosis and rehabilitation utilising computational tools~\cite{ringeval2019avec}. Head motion patterns have nevertheless received little attention. Psychological research on depression assessment has identified head motion as a significant non-verbal cue for depression with more pronounced behavioural changes in hand and head regions as compared to other body parts for depressed patients \cite{pedersen1988ethological}. Waxer \emph{et. al.}~\cite{waxer1974nonverbal} found that depressed subjects are more likely to keep their heads in a downward position and exhibit significantly reduced head nodding compared to healthy subjects \cite{fossi1984ethological}. Another study focusing on social interactions identified the reduced involvement of depressed patients in conversations, where their behaviour was characterised by lesser encouragement (head nodding and backchanneling while listening) and fewer head movements~\cite{hale1997non}.

From a computational standpoint, only a few studies have employed head pose and movement patterns for automatic depression detection. Alghowinem~\emph{et al.}~\cite{alghowinem2013head} analysed head movements by modelling statistical features extracted from the 2D Active Appearance Model (AAM) projection of a 3D face and demonstrated the efficacy of head pose as behavioural cue. Another study~\cite{joshi2013can} generated a histogram of head movements normalised over time to highlight the diminished movements of depressed patients due to psychomotor retardation, characterised by a more frequent occurrence of static head positions than in healthy controls. Several studies \cite{song2020spectral, dibekliouglu2017dynamic,cohn2018multimodal, morales2017cross} explored the utilisation of head motion as a complementary cue to other modalities to enhance detection performance. For instance, several studies~\cite{alghowinem2016multimodal, alghowinem2020interpretation} combined head pose with speech behaviour and eye gaze to develop statistical features for depression analysis. Generalisation across different cross-cultural datasets was attempted in~\cite{alghowinem2015cross} by using head pose and eye gaze based temporal features. Kacem~\emph{et. al.}~\cite{kacem2018detecting} encoded head motion dynamics with facial expressions to classify depression based on severity, while Dibeklioglu \emph{et. al.}~\cite{dibekliouglu2015multimodal} included vocal prosody in combination with head and facial movements for depression detection. 
%

%
%
\subsection{Novelty of the Proposed Approach} 
From the literature review, it can be seen that while a number of studies have employed head movements as a complementary cue in multimodal approaches, only few studies have deeply explored head motion as a rich source of information. Further, the explainability of behavioural features, especially head motion features, for depression detection has not yet been explored in the literature. This study (a) is the first to propose the use of kinemes as depression biomarkers, (b) explores multimodal cues derived from head motion behaviour as potential biomarkers for depression; specifically, we show that kinemes learned for the depressed and control classes, or only the control class enable accurate depression detection, and (c) the learned kinemes also \textit{explain} depressed behaviours consistent with prior observations.


\section{Kineme Formulation}\label{Sec:KF}

This section describes our approach to discovering a set of elementary head motion units termed \emph{kinemes} from 3D head pose angles. These head pose angles are expressed as a time-series of short overlapping segments, which enables shift invariance. The segments are then projected onto a lower-dimensional space and clustered using a Gaussian Mixture Model \cite{samanta2017role}. 

We extracted 3D head pose angles using the OpenFace tool~\cite{Baltrusaitis16} in terms of 3D Euler rotation angles, \emph{pitch} ($\theta_p$), \emph{yaw} ($\theta_y$) and \emph{roll} ($\theta_r$). The head movement over a duration $T$ is denoted as a time-series: $\boldsymbol{\theta} = \{\theta_p^{1:T}, \theta_y^{1:T}, \theta_r^{1:T}\}$. We ensure that the rotation angles remain non-negative by defining the range in [0$^{\circ}$, 360$^{\circ}$]. 

For each video, the multivariate time-series $\boldsymbol{\theta}$ is divided into short overlapping segments of length $\ell$ with overlap $\ell/2$, where the $i^{th}$ segment is represented as a vector $\mathbf{h}^{(i)} = [\theta_p^{i:i+\ell}\, \theta_y^{i:i+\ell}\, \theta_r^{i:i+\ell}]$. Considering the total number of segments in any given video as $s$, the characterisation matrix  $\mathbf{H}_{\boldsymbol\theta}$ for this video is defined as
$ \mathbf{H}_{\boldsymbol\theta} = [\mathbf{h}^{(1)}, \mathbf{h}^{(2)},\cdots, \mathbf{h}^{(s)}]$. Thus, for a training set of $N$ samples, the head motion matrix is created as $\mathbf{H} = [\mathbf{H}_{\boldsymbol\theta_1}|\mathbf{H}_{\boldsymbol\theta_2}|\cdots|\mathbf{H}_{\boldsymbol\theta_N}]$ with each column of $\mathbf{H}$ representing a single head motion time-series for a given video sample. We decompose $\mathbf{H}\in\mathbb{R}_+^{m\times n}$ into a basis matrix $\mathbf{B}\in\mathbb{R}_+^{m\times q}$ and a coefficient matrix $\mathbf{C}\in\mathbb{R}_+^{q\times n}$ using Non-negative Matrix Factorization (NMF) such that $m = 3\ell$, $n = Ns$
\vspace{-1mm}
\begin{equation}
    \underset{\mathbf{B} \geq 0, \mathbf{C} \geq 0}{\text{ min}} \lVert{\mathbf{H} - \mathbf{B}\mathbf{C}}\rVert_F^2
\end{equation}
where $q \leq min(m, n)$ and $\lVert \textbf{ . }  \rVert_F$ denotes the Frobenius norm. Rather than clustering the raw head motion segments, we employ a more interpretable and stable approach by clustering the coefficient vectors in the transformed space. To this end, we learn a Gaussian Mixture Model (GMM) using the columns of the coefficient matrix $\mathbf{C}$ to produce a ${\mathbf{C}^*}\in\mathbb{R}_+^{q\times k}$ where $k << Ns$. These vectors in the learned subspace are transformed back to the original head motion subspace defined by the Euler angles using $\mathbf{H}^*=\mathbf{B}\mathbf{C}^*$. The columns of matrix $\mathbf{H}^*$ represent the set of $K$ kinemes as $\{\mathcal{K}_i\}_{i=1}^K$. 

Now, we can represent any head motion time-series $\theta$ as a sequence of kinemes discovered from the input video set by associating each segment of length $\ell$ from $\theta$ with one of the kinemes. For each $i^{th}$ segment in the time-series, we compute the characterisation vector $\mathbf{h}^{(i)}$ and project it onto the transformed subspace defined by $\mathbf{B}$ to yield $\mathbf{c}^{(i)}$ such that:\vspace{-1mm}
\begin{equation}
    \hat{\mathbf{c}} = \underset{\mathbf{c}^{(i)} \geq 0}{\text{arg min}} \lVert{\mathbf{h}^{(i)} - \mathbf{B}\mathbf{c}^{(i)}}\rVert_F^2
\end{equation}
We then maximise the posterior probability $P({K}|\hat{\mathbf{c}})$ over all kinemes to map the $i^{th}$ segment with its corresponding kineme $K^{(i)}$. In the same way, we compute the corresponding kineme label for each segment of length $\ell$ to obtain a sequence of kinemes: $\{K^{(1)} \cdots K^{(s)}\}$, where $K^{(j)}\in \mathcal{K}$ for all segments of time-series $\theta$. 

\section{Explainable Kineme Features }\label{Sec:EKF}

%
%
\begin{figure*}[t]
\centering
  \includegraphics[width=0.35\linewidth]{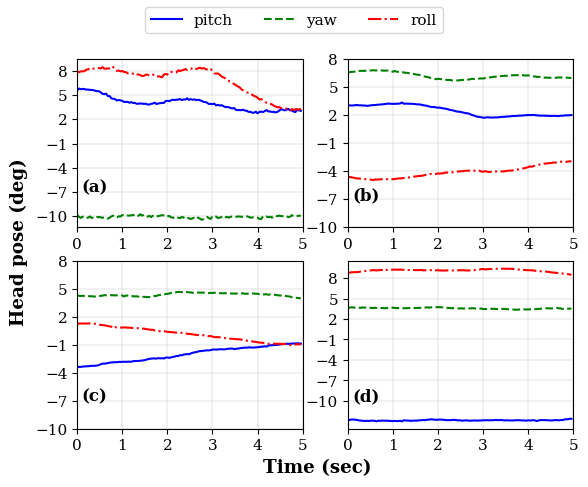}\hspace{0.25cm}\includegraphics[width=0.35\linewidth]{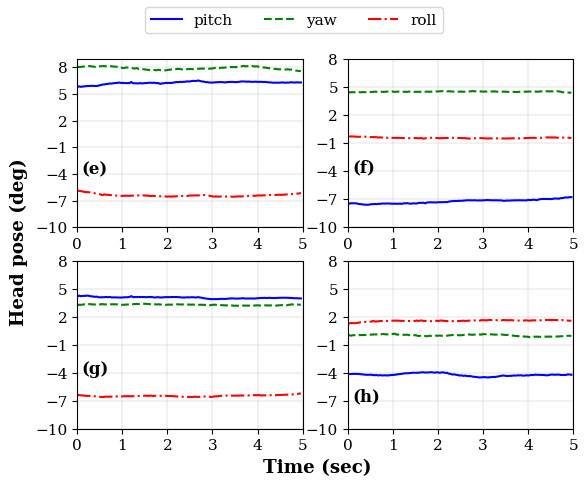}
        \vspace{-4mm}
\caption{Plots of kinemes that occur more frequently for the control (left) and patient (right) cohorts in the \textit{BlackDog} dataset.}\label{fig:kinemes_bdi}\vspace{-4mm}
\end{figure*}
\begin{figure*}[t]
\centering
    \includegraphics[width=0.35\linewidth]{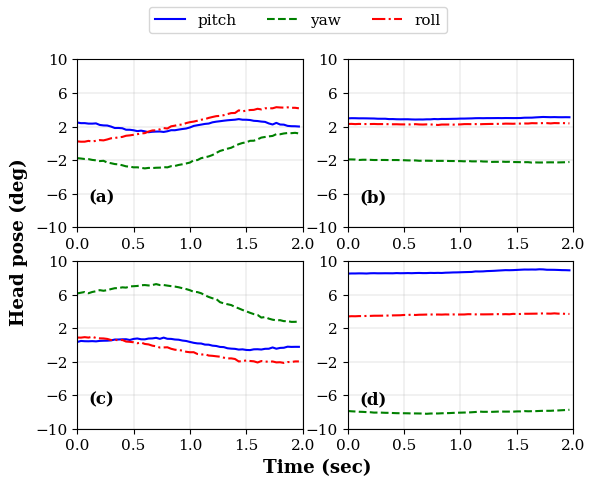}\hspace{0.25cm}\includegraphics[width=0.35\linewidth]{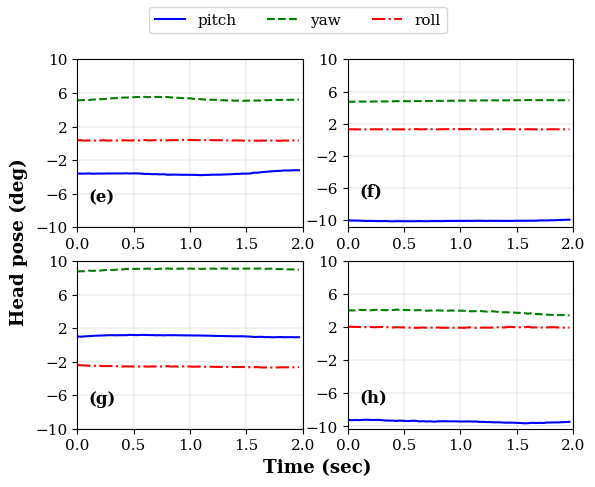}
        \vspace{-4mm}
\caption{Plots of kinemes that occur more frequently for the minimally depressed (left) and patient (right) cohorts in \textit{AVEC2013}.}\label{fig:kinemes_avec}\vspace{-4mm}
\end{figure*}


We now examine kineme patterns obtained from the depression datasets, namely \textit{BlackDog}~\cite{alghowinem2016multimodal} and \textit{AVEC2013}~\cite{valstar2013avec} (described in Sec. ~\ref{sec:datasets}). Using the \textit{Openface}~\cite{Baltrusaitis16} toolkit, we extracted \textit{yaw}, \textit{pitch} and \textit{roll} angles per frame, and segmented each video into 2s and 5s-long chunks with 50\% overlap for the AVEC2013 and BlackDog datasets, respectively. Considering $K = 16$ \cite{samanta2017role}, we extracted kinemes from both patient and healthy control segments, following the procedure outlined in Sec.~\ref{Sec:KF}. We further examined the kinemes learned for each dataset to identify the set of distinctive kinemes for the two classes. To obtain the most discriminative kinemes, we computed the relative frequency of occurrence for each kineme for the control and patient data, and selected the top five kinemes per class based on their relative frequency difference (see Sec.~\ref{Sec:approach1}).  

Selected kinemes corresponding to the maximal difference in their relative frequency of occurrence for the control and patient classes are visualised in Figures~\ref{fig:kinemes_bdi} (\textit{BlackDog}) and~\ref{fig:kinemes_avec} (\textit{AVEC2013}). Examining the control-specific kinemes in Figs.~\ref{fig:kinemes_bdi} and~\ref{fig:kinemes_avec}, we observe a greater degree of movement for healthy subjects as compared to a predominantly static head pose conveyed by the depressed patient-specific kinemes. Head nodding, characterised by pitch oscillations, and considerable roll angle variations can be noted for at least one control-class kineme; conversely, patient-specific kinemes exhibit relatively small changes over all head pose angular dimensions. These findings are reflective of reduced head movements in the depressed cohort compared to healthy individuals, which is consistent with observations made in past studies \cite{hale1997non, alghowinem2013head}. 

\section{Classification Methodology}\label{Sec:Meth}

In this section, we outline the methodology for discovering kinemes from short overlapping video segments. Initially, we discovered kinemes utilising data segments from both control and patient classes (two-class kineme discovery or 2CKD approach; see Section~\ref{Sec:EKF}). Subsequently, we learned kinemes solely from the healthy control cohort and utilised them to represent the head pose data of depressed patients (denoted as healthy control kineme discovery or HCKD) approach. 

Given a time series $\boldsymbol{\theta}$, we divide it into short overlapping segments of uniform length, with a segment duration of $5s$ for the BlackDog and $2s$ for the AVEC dataset. These segment lengths were empirically chosen and provided the best results from among segment lengths spanning $2s$ to $7s$ for both datasets. For both approaches, a total of $K = 16$ kinemes are learned from the two datasets as per the procedure outlined in Section~\ref{Sec:KF}.

%
%
\subsection{Kineme Discovery from Two-class Data} \label{Sec:approach1}
To examine whether the kinemes discovered from head pose angles of both classes are effective cues for depression detection, we learn kinemes from segments corresponding to both patient and control videos. 
Upon discovering the kineme values, the \emph{relative frequency} $\eta_{K_i}$ of each kineme $K_i$ is computed over the two classes as:
\vspace{-1mm}
\begin{equation}
    \mathbf{\eta}_{K_i} = \frac{\emph{f } ( K_i) } {\sum_{i=1}^{16} \emph{f } ( K_i)}
\end{equation}
where $\emph{f} (K_i)$ represents the frequency of occurrence of the kineme $K_i$ for a particular class. We then compute the relative frequency difference for each kineme between the two classes to identify the ten most differentiating kinemes (four kinemes per class are depicted in Figs.~\ref{fig:kinemes_bdi}, \ref{fig:kinemes_avec}). Next, we generate a feature set by extracting the frequencies of the selected kinemes over the thin-slice chunks considered for analysis. Thus, we obtain a 10-dimensional feature vector representing kineme frequencies for each chunk.

%
%
\subsection{Kineme Discovery from Control Data}
\label{Sec:approach2}
Here, we learn kinemes representing head motion solely from the control cohort. Subsequently, head pose segments from both the patient and control classes are represented via the discovered kinemes, and reconstruction errors computed.  Let the raw head pose vector $\mathbf{h}^{(i)}$ for the $i^{th}$ segment in the original subspace be denoted as:
\begin{equation}
\mathbf{h}^{(i)} = [\theta_p^{i:i+\ell}\, \theta_y^{i:i+\ell}\, \theta_r^{i:i+\ell}]
\end{equation}
Let the kineme value associated with this segment be $K^{(i)}$. Based on the kinemes discovered from the control cohort alone, we calculate the reconstructed kineme for the $i^{th}$ segment as $\tilde{\mathbf{h}}^{(i)}$. The reconstructed vector for each kineme is determined by converting the GMM cluster centre for each kineme from the learned space to the original \emph{pitch}-\emph{yaw}-\emph{roll} space. The reconstructed head pose vector for the segment is:
\begin{equation}
    \tilde{\mathbf{h}}^{(i)} = [\tilde{\theta}_p^{i:i+\ell}\, \tilde{\theta}_y^{i:i+\ell}\, \tilde{\theta}_r^{i:i+\ell}]
\end{equation}
To compute the reconstruction error for both depressed patients and healthy controls, we compute the signed difference between the two vectors for each segment to account for the difference between raw head pose vector and the GMM cluster centres.
We calculate the difference vector $\mathbf{d}^{(i)}$ for each $i^{th}$ segment as: 
\vspace{-1mm}
\begin{equation}
    \mathbf{d}^{(i)} = \mathbf{h}^{(i)} - \tilde{\mathbf{h}}^{(i)} = [d_p^{i:i+\ell}\, d_y^{i:i+\ell}\, d_r^{i:i+\ell}]
\end{equation}
These signed differences values are added over each angular dimension of pitch (\emph{p}) , yaw (\emph{y}), and roll (\emph{r}) for the segment.
\begin{equation}
        {s_e}^{(i)} = \sum_{n=1}^{\ell} d_e^{i:i+n} 
\end{equation}
where each ${s_e}^{(i)}$ is calculated for each angular dimension $e \in \{p, y, r\}$ over all segments of both classes. 
Depending on the thin-slice chunk duration considered for classification, we compute different descriptive statistics to generate the feature set. Considering number of elementary kineme chunks in the considered time-window to be $n_c$, we obtain the following feature vector for each angle  $e \in \{p, y, r\}$:
\vspace{-1mm}
\begin{equation}
        \mathbf{as_e} = [\lvert{s_e}^{(1)} \rvert, \lvert{s_e}^{(2)}\rvert, \cdots, \lvert{s_e}^{(n_c)}\rvert ] 
\end{equation}
where $\lvert \cdot \rvert$ represents the absolute value. We then calculate eight statistical features from the above vectors, namely, \emph{minimum}, \emph{maximum}, \emph{range}, \emph{mean}, \emph{median}, \emph{standard deviation}, \emph{skewness}, and \emph{kurtosis} (total of $8 \times 3$ features over the yaw, pitch, roll dimensions).

\section{Experiments }\label{Sec:ER}

%
%
\begin{sloppypar}
We perform binary classification on the BlackDog and AVEC2013 datasets, plus 4-class classification on AVEC2013. This section details our datasets, experimental settings and learning algorithms.
\end{sloppypar}

%
%
\subsection{Datasets}
\label{sec:datasets}
We examine two datasets in this study: clinically validated data collected at the Black Dog Institute -- a clinical research facility focusing on the diagnosis and treatment of mood disorders such as anxiety and depression (referred to as \emph{BlackDog} dataset) -- and the \emph{Audio/Visual Emotion Challenge} (AVEC2013) depression dataset. 

\textbf{BlackDog Dataset~\cite{alghowinem2016multimodal}:} This dataset comprises responses from healthy controls and depression patients selected as per the criteria outlined in the Diagnostic and Statistic Manual of Mental Disorders (DSM-IV). Healthy controls with no history of mental illness and patients diagnosed with severe depression were carefully selected~\cite{alghowinem2016multimodal}. For our analysis, we focus on the structured interview responses in~\cite{alghowinem2016multimodal}, where participants answered open-ended questions about life events, designed to elicit spontaneous self-directed responses, asked by a clinician. 
In this study, we analyse video data from 60 subjects (30 depressed patients and 30 healthy controls), with interview durations ranging from $183-1200s$. 

\textbf{AVEC2013 Dataset~\cite{valstar2013avec}:} Introduced for a challenge in 2013, this dataset is a subset of the audio-visual depressive language corpus (AViD-corpus) comprising 340 video recordings of participants performing different PowerPoint guided tasks detailed in~\cite{valstar2013avec}. The videos are divided into three nearly equal partitions (training, development, and test) with videos ranging from $20-50 min$. Each video frame depicts only one subject, although some participants feature in multiple video clips. The participants completed a multiple-choice inventory based on the Beck Depression Index (BDI)~\cite{beck1996comparison} with scores ranging from 0 to 63 denoting the severity of depression. For binary classification, we dichotomise the recordings into the non-depressed and depressed cohorts as per the BDI scores. Subjects with a BDI score $\leq 13$ are categorised as \emph{non-depressed}, while the others are considered as \emph{depressed}. 

\textbf{AVEC2013 Multi-Class Classification:} For fine-grained depression detection over the AVEC2013 dataset, we categorise the dataset based on the BDI score into four classes as detailed below: 
\begin{itemize}
    \item Nil or minimal depression: BDI score 0 - 13
    \item Mild depression: BDI score 14 - 19
    \item Moderate depression: BDI score 20 - 28
    \item Severe depression: BDI score 29 - 63
\end{itemize}

%
%
\subsection{Experimental Settings}
\textbf{Implementation Details:} For binary classification, we evaluate performance for the smaller \emph{BlackDog} dataset via 5-repetitions of 10-fold cross-validation (10FCV). For the AVEC2013, the pre-partitioned train, validation and test sets are employed. We utilise the validation sets for fine-tuning classifier hyperparameters.

\textbf{Chunk vs Video-level Classification:} The videos from both datasets are segmented into smaller chunks of $15s - 135$s length, to examine the influence of \emph{thin-slice} chunk duration on the classifier performance. We repeated the video label for all chunks and metrics are computed over all chunks for chunk-level analysis. Additionally, video-level classification results are obtained by computing the majority label over all video chunks in the test set. 

\textbf{Performance Measures:} For the BlackDog dataset, results are shown as $\mu \pm \sigma$ values over 50 runs ($5\times$ 10FCV repetitions). For AVEC2013, performance on the test set is reported. For both, we evaluate performance via the accuracy (Acc), weighted F1 (F1), precision (Pr), and recall (Re) metrics. The weighted F1-score denotes the mean F1-score over the two classes, weighted by class size.

%

%
%
\subsection{Classification Methods}
Given that our proposed features do not model spatial or temporal correlations, we employ different machine learning models for detecting depression as described below:
\begin{itemize}
    \item \textbf{Logistic Regression (LR)}, a probabilistic classifier that employs a sigmoid function to map input observations to binary labels. We utilise extensive grid-search to fine-tune parameters such as penalty $\in \{ l1, l2, None \}$ and regulariser $\lambda \in \{ 1e^{-6}, \cdots, 1e^{3}\}$.
    \item\textbf{Random Forest (RF)}, where multiple decision trees are generated from training data whose predictions are aggregated for labelling. Fine-tuned parameters include the number of estimators $N \in [2, \cdots,  8]$, maximum depth $\in [3, \cdots,  7]$, and maximum features in split $\in [3, \cdots,  7]$.
    \item \textbf{Support Vector Classifier (SVC)}, a discriminative classifier that works by transforming training data to a high-dimensional space where the two classes can be linearly separated via a hyperplane. For SVC, we examine different kernels $\in \{ rbf, poly, sigmoid\}$ and fine-tune regularisation parameter $C \in \{ 0.1, 1, 10, 100 \}$ and kernel coefficient $\gamma \in \{ 0.0001, \cdots, 1, scale, auto \}$. 
    \item \textbf{Extreme Gradient Boosting (XGB)}, a model built upon a gradient boosting framework, and focused on improving a series of weak learners by employing the gradient descent algorithm in a sequential manner. The fine-tuned hyperparameters include the number of estimators $\{ 50, 100, 150 \}$, maximum depth $\in [3, \cdots,  7]$ of the tree and learning rate $\in [0.0005, \cdots,  0.1]$.
    \item \textbf{Multi Layer Perceptron (MLP)}, where we employed a feed-forward neural network with two hidden dense layers comprising 12 and 6 neurons, resp., with a rectified linear unit (ReLU) activation. For training, we employ categorical cross-entropy as the loss function and fine-tune the following hyperparameters: learning rate $\in \{ 1e^{-4}, 1e^{-3}, 1e^{-2}\}$, and batch size $\in \{ 16, 24, 32, 64 \}$. We utilise the Adam optimiser for updating the network weights during training.
\end{itemize}

%
%
\section{Results and Discussion}\label{sec:ResultsDiscussion}

\begin{table*}[ht]
 \caption{Chunk and Video-level classification results on the BlackDog dataset with the 2CKD and HCKD approaches. Accuracy (Acc), F1, Precision (Pr) and Recall (Re) are tabulated as ($\mu \pm \sigma$) values. } \vspace{-2mm}


\begin{tabular}{|l l||cc cc||cc cc|} 
\hline
    \bf Condition & \bf Classifier    & \multicolumn{4}{c||}{\textbf{Chunk-level}} & \multicolumn{4}{c|}{\textbf{Video-level}}    \\ 
     & & \textbf{Acc} & \textbf{F1} & \textbf{Pr} & \textbf{Re} & \textbf{Acc} & \textbf{F1} & \textbf{Pr} & \textbf{Re}\\ 
    \hline\hline
    \multirow{5}{*}{\bf 2CKD}& \textbf{LR} &  0.60±0.15 & 0.61±0.14 &  0.67±0.22 & 0.65±0.22 & 0.60±0.20 & 0.59±0.21 &  0.55±0.30 & 0.65±0.33  \\ 
    & \textbf{RF} & 0.58±0.13 & 0.60±0.12 &  0.67±0.21 & 0.61±0.21  &0.61±0.19 & 0.62±0.19 &  0.59±0.32 & 0.59±0.32 \\ 
    & \textbf{SVC} & 0.60±0.15 & 0.62±0.15&  0.68±0.25 & 0.62±0.25  & 0.62±0.19 & 0.63±0.19 &  0.61±0.32 & 0.59±0.33 \\
    & \textbf{XGB}& 0.55±0.17 & 0.54±0.16 &  0.63±0.21 & 0.71±0.22  & 0.53±0.17 & 0.50±0.20 &  0.54±0.23 & 0.79±0.21 \\ 
    & \textbf{MLP} & 0.53±0.15 & 0.52±0.17 &  0.60±0.22 & 0.71±0.21 & 0.51±0.20 & 0.47±0.21 &  0.53±0.27 & 0.74±0.32 \\ \hline
    \multirow{5}{*}{\bf HCKD} & \textbf{LR}  & 0.77±0.13 & 0.78±0.12 &  0.85±0.19 & 0.74±0.21 & 0.79±0.16 & 0.78±0.17 &  0.81±0.30 & 0.66±0.31 \\ 
    & \textbf{RF} &0.71±0.13 & 0.73±0.12&  0.75±0.25 & 0.71±0.20  &0.76±0.15 & 0.76±0.16 &  0.75±0.26 & 0.82±0.25 \\ 
    & \textbf{SVC} & {0.78±0.14} & \textbf{0.79±0.13}&  0.87±0.18 & 0.74±0.20  & {0.80±0.18} & \textbf{0.80±0.19} &  0.83±0.30 & 0.70±0.31 \\ 
    & \textbf{XGB}& 0.72±0.13 & 0.72±0.12&  0.75±0.18 & 0.81±0.15  & 0.78±0.17 & 0.78±0.17 &  0.74±0.27 & 0.82±0.27  \\ 
    & \textbf{MLP} & 0.75±0.13 & 0.76±0.12 &  0.78±0.21 & 0.81±0.21 & 0.76±0.16 & 0.76±0.16 &  0.74±0.29 & 0.77±0.28 \\
    \hline
\end{tabular} 
\vspace{-2mm}
\label{tab:BDI_res1}
\end{table*}
%
%
\begin{table*}[ht]
\caption{\label{tab:AVEC_res1} Chunk and Video-level classification results on the AVEC2013 dataset with the 2CKD and HCKD approaches. Accuracy (Acc), F1, Precision (Pr) and Recall (Re) are tabulated as ($\mu \pm \sigma$) values.}
\begin{tabular}{|l l||cc cc||cc cc|} 
    \hline
    \bf Condition & \bf Classifier    & \multicolumn{4}{c||}{\textbf{Chunk-level}} & \multicolumn{4}{c|}{\textbf{Video-level}}    \\ 
     & & \textbf{Acc} & \textbf{F1} & \textbf{Pr} & \textbf{Re} & \textbf{Acc} & \textbf{F1} & \textbf{Pr} & \textbf{Re}\\ 
    \hline\hline

   \multirow{5}{*}{\bf 2CKD}&  \textbf{LR} & 0.58 & 0.58  & 0.54 & 0.65 & 0.61 & 0.61 & 0.57 & 0.71   \\ 
    & \textbf{RF} & 0.61 & 0.61 & 0.57 & 0.59 & 0.72 & 0.72 & 0.67 & 0.82 \\ 
    & \textbf{SVC} & 0.61 & 0.61 & 0.57 & 0.63  & 0.64 & 0.64 & 0.61 & 0.65 \\ 
    & \textbf{XGB} & 0.59 & 0.58 & 0.57 & 0.44 & 0.67 & 0.67 & 0.65 & 0.65 \\ 
    & \textbf{MLP}    & 0.56 & 0.56 & 0.52 & 0.60 & 0.58 & 0.58 & 0.65 & 0.65 \\ \hline
    \multirow{5}{*}{\bf HCKD} & \textbf{LR} & 0.80 & 0.80 & 0.77 & 0.81  & 0.94  & 0.94 & 0.94  & 0.94\\ 
    & \textbf{RF} & 0.78 & 0.78 & 0.78 & 0.75   & {1.00} & \textbf{1.00} & 1.00 & 1.00 \\ 
    & \textbf{SVC} & {0.82} & \textbf{0.82} & 0.83 & 0.77  &  {1.00} & \textbf{1.00} & 1.00 & 1.00 \\ 
    & \textbf{XGB} & 0.80 & 0.80  & 0.79 & 0.77 & {1.00} & \textbf{1.00} & 1.00 & 1.00\\ 
    & \textbf{MLP}    & 0.81 & 0.80 & 0.80 & 0.77 & {1.00} & \textbf{1.00} & 1.00 & 1.00\\
    \hline
\end{tabular}
\end{table*}
%
\begin{table*}[ht]
 \caption{Comparison with prior works for the two datasets.} \vspace{-2mm}
\begin{tabular}{|l l l||cc cc|} 
\hline
    \bf Dataset & \bf Methods  & \bf Features    & \multicolumn{4}{c|}{\textbf{Evaluation metrics}}     \\ 
     & & &\textbf{Acc} & \textbf{F1} & \textbf{Pr} & \textbf{Re}\\ 
    \hline\hline
    \multirow{4}{*} {\textbf{BlackDog}} & Alghowinem \etal \ \cite{alghowinem2013head} & Head movement &  - & - &  - & 0.71  \\ 
     & Joshi \etal \ \cite{joshi2013can} & Head movement &  0.72 & - &  - & -    \\ 
     & Ours (Chunk-level) & Kinemes &  \textbf{0.75} & {0.76} & { 0.78 }& \textbf{0.81}    \\
    & Ours (Video-level) & Kinemes &  \textbf{0.80} & {0.80} & { 0.83} &{0.70}  \\  \hline
    \multirow{4}{*}{\textbf{AVEC2013}}  & Senoussaoui \etal \ (AVEC2014)~\cite{senoussaoui2014model} & Video features &  0.82 & - &  - & - \\ 
    & Al-gawwam~\etal \ (AVEC2014 - Northwind) \cite{al2018depression} & Eye Blink &  0.85 & - &  - & -    \\ 
    & Al-gawwam~\etal \ (AVEC2014 - Freeform) \cite{al2018depression} & Eye Blink &  0.92 & - &  - & -    \\ 
     & Ours (AVEC2013 at chunk-level) & Kinemes &  0.82 & 0.82 &  0.83 & 0.87    \\
    & Ours (AVEC2013 at Video-level) & Kinemes &  \textbf{1.00} & {1.00} & {1.00} & {1.00}\\  \hline
\end{tabular} 
\vspace{-2mm}
\label{tab:Comp}
\end{table*}

\begin{table*}[ht]
\caption{\label{tab:AVEC_res_4class} Video-level 4-class categorization results on the AVEC dataset obtained with the HCKD approach. Accuracy (Acc), F1, Precision (Pre) and Recall (Re) are tabulated.} 
\vspace{-2mm}
\begin{tabular}{|l l||cc cc|} 
    \hline
    \bf Condition & \bf Classifier     & \multicolumn{4}{c|}{\textbf{Video-level}} \\ 
               & & \textbf{Acc} & \textbf{F1} & \textbf{Pr} & \textbf{Re} \\ 
    \hline\hline

    \multirow{5}{*}{\bf HCKD} & \textbf{LR}  & 0.71 & \textbf{0.72} & 0.73 & 0.71 \\ 
    & \textbf{RF}  & {0.74} & \textbf{0.72}  & 0.80 & 0.74 \\ 
    &  \textbf{SVC} & {0.74} & \textbf{0.72} & 0.75 & 0.74 \\ 
    &  \textbf{XGB}  & 0.71 & 0.69  & 0.68 & 0.71 \\ 
    &  \textbf{MLP}  & 0.69 & 0.66  & 0.64 & 0.69 \\
    \hline
\end{tabular}\vspace{-2mm}
\end{table*}


Table ~\ref{tab:BDI_res1} shows the classification results obtained for the \emph{BlackDog} dataset with the 2CKD and HCKD approaches (Section~\ref{Sec:Meth}). Table ~\ref{tab:AVEC_res1} presents the corresponding results for the \emph{AVEC2013} dataset. These tables present classification measures obtained at the \emph{chunk-level} (best results achieved over $15 - 135s$-long chunks for the two datasets are presented), and the \emph{video-level} (label derived upon computing the mode over the chunk-level labels). Based on these results, we make the following observations:
\begin{itemize}
    \item It can be noted from Tables~\ref{tab:BDI_res1} and~\ref{tab:AVEC_res1} that relatively lower accuracies and F1 scores are achieved for both datasets using the 2CKD approach, implying that while class-characteristic kinemes are explanative as seen from Figs.~\ref{fig:kinemes_bdi} and~\ref{fig:kinemes_avec}, they are nevertheless not discriminative enough to effectively distinguish between the two classes. 
    \item In comparison, we note far superior performance with the HCKD method over all classifiers. As a case in point, we obtaine peak chunk-level F1-scores of 0.79 and 0.62, resp., for HCKD and 2CKD on BlackDog, while the corresponding F1-scores are 0.82 and 0.61, resp., on AVEC. This observation reveals considerable and distinguishable differences in the reconstruction errors for the patient and control classes, and convey that patient data are characterised as \emph{anomalies} when kinemes are only learned from the control cohort.
    \item Examining the HCKD precision and recall measures for both datasets, we note higher precision than recall at the chunk-level for the BlackDog dataset. Nevertheless, higher recall is achieved at the video-level with multiple classifiers. Likewise, higher chunk-level precision is noted for AVEC, even if ceiling video-level precision and recall are achieved.
    \item Comparing HCKD chunk and video-level F1-scores for both datasets, similar or higher video-level F1 values can be seen in Table~\ref{tab:BDI_res1}. F1-score differences are starker in Table \ref{tab:AVEC_res1}, where video-level scores are considerably higher than chunk-level scores. These results suggest that aggregating observations over multiple thin-slice chunks is beneficial and enables more precise predictions as shown in~\cite{madan_gahalawat_guha_subramanian_ICMI2021_Kinemes}.
    \item Examining measures achieved with the different classifiers, the support vector classifier achieves the best chunk-level F1-score on both datasets, with the LR classifier performing very comparably. All classifiers achieve very similar performance when video-level labels are compared.
\end{itemize}
%

%
%
\subsection{Comparison with the state-of-the-art} Our best results are compared against prior classification-based depression detection studies in Table~\ref{tab:Comp}. For the BlackDog dataset, Alghowinem \emph{et al.}~\cite{alghowinem2013head} analysed statistical functional features extracted from a 2D Active Appearance Model, whereas Joshi \emph{et al.}~\cite{joshi2013can} computed a histogram of head movements by estimating the displacement of fiducial facial points. Compared to \textit{N}-average recall of 0.71 reported in~\cite{alghowinem2013head}, and an accuracy of 0.72 noted in~\cite{joshi2013can}, our kineme-based approach achieves better chunk and video-level accuracies (0.75 and 0.80, resp.), and  superior chunk-level recall (0.81). As most previous studies on the AVEC2013 dataset focus on continuous prediction, we compare our model's performance with the AVEC2014~\cite{valstar2014avec} results examining visual features. 

AVEC2014 used the same subjects as AVEC2013, but with additional, specific task data (\emph{Northwind}, \emph{Freeform}) extracted from the AViD videos. For video analysis, Senoussaoui \etal~\cite{senoussaoui2014model} extracted LGBP-TOP features from frame blocks to obtain an accuracy of 0.82 using an SVM classifier. On the other hand, Al-gawwam \etal~\cite{al2018depression} extracted eye-blink features from video data using a facial landmark tracker to achieve an accuracy of 0.92 for the \emph{Northwind} task and 0.88 for the \emph{Freeform} task. Comparatively, our work achieves an accuracy of 0.82 at the chunk-level and 1.00 at the video-level. The next section will detail the performance of a more fine-grained 4-class categorisation on the AVEC2013 dataset.

%
%
\subsection{\textbf{AVEC2013 Multi-class Classification}} 
\label{sec: avec_4class_res}
Table~\ref{tab:AVEC_res_4class} depicts video-level 4-class classification results achieved on the AVEC2013 dataset via the HCKD approach. The 4-class categorisation was performed to further validate the correctness of the HCKD approach, which produces ceiling video-level F1, Precision and Recall measures on AVEC2013 in binary classification. Results are reported on the test set, upon fine-tuning the classifier models on the development set. Reasonably good F1-scores are achieved even with 4-class classification, with a peak F1 of 0.72 obtained with the LR, RF and support vector classifiers. Cumulatively, our empirical results confirm that kinemes encoding atomic head movements are able to effectively differentiate between (a) the patient and control classes, and (b) different depression severity bands. 

\begin{figure*}[t]
\centering
    \includegraphics[width=0.45\linewidth]{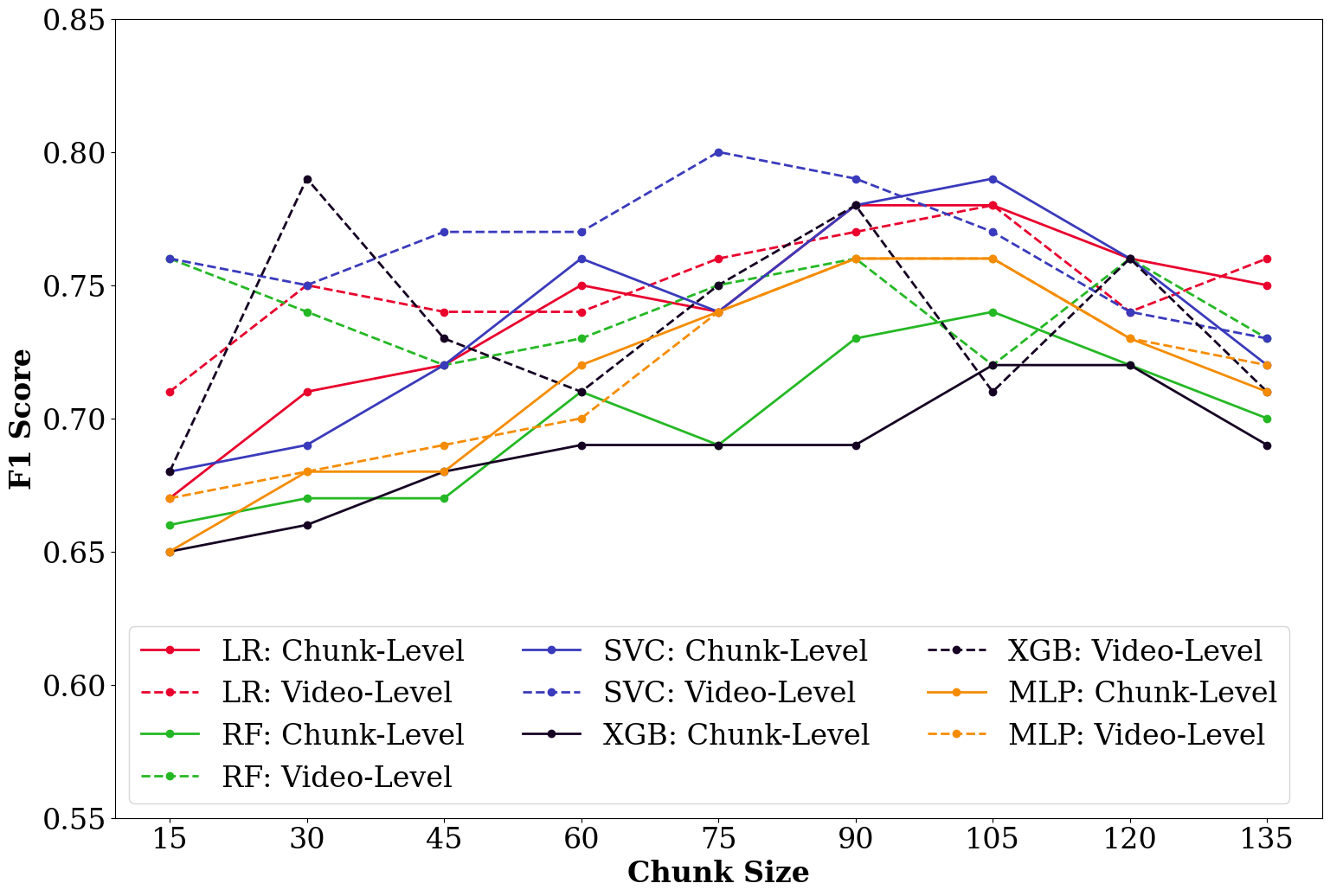}\hspace{0.15cm}\includegraphics[width=0.45\linewidth]{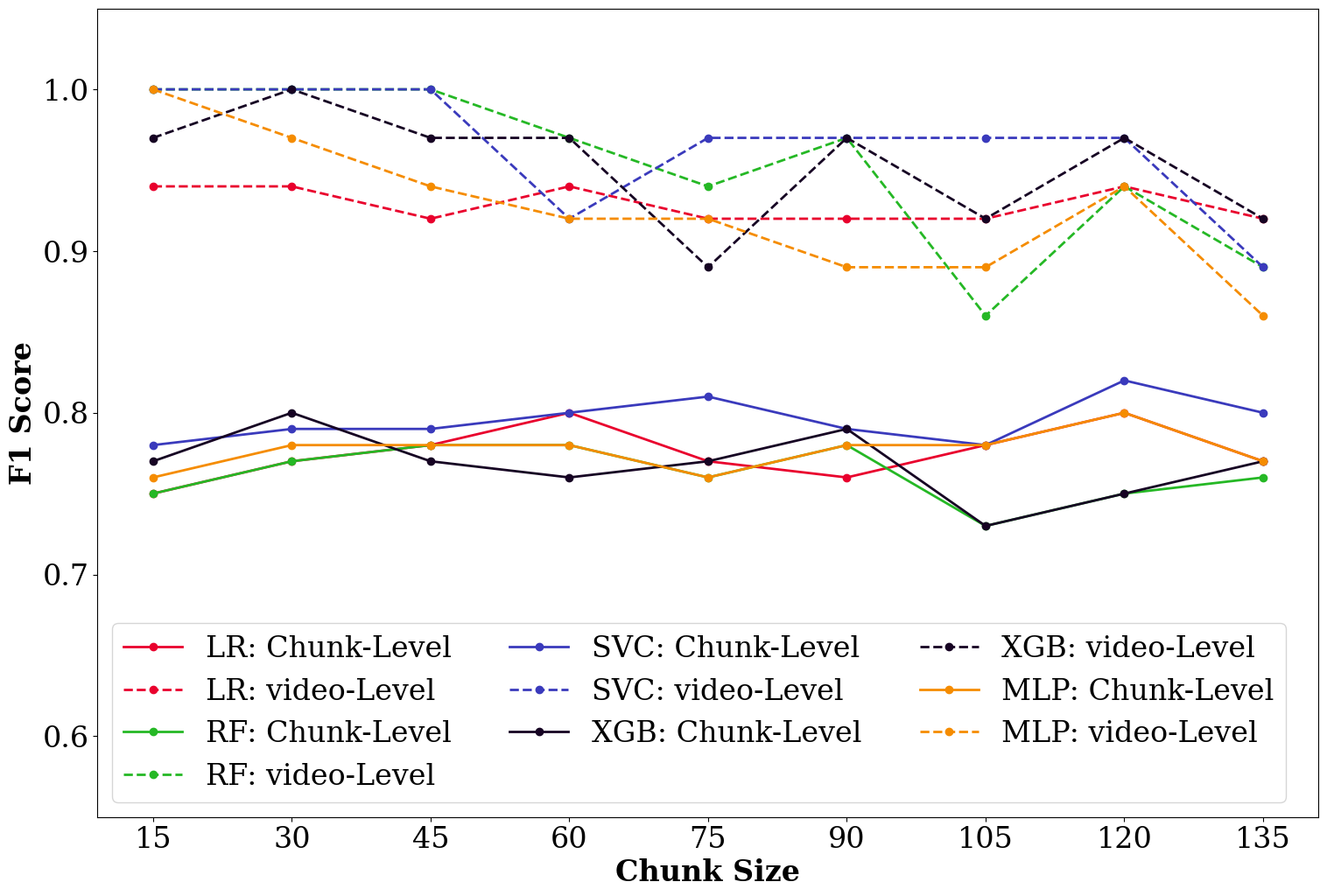}
        \vspace{-2mm}\caption{Chunk vs video-level performance comparison for the BlackDog (left) and AVEC2013 (right) datasets.}\label{fig:Chunk_vs_Vid_kineme}
\end{figure*}

%
%

\begin{figure*}[t]
\centering
    \includegraphics[width=0.45\linewidth]{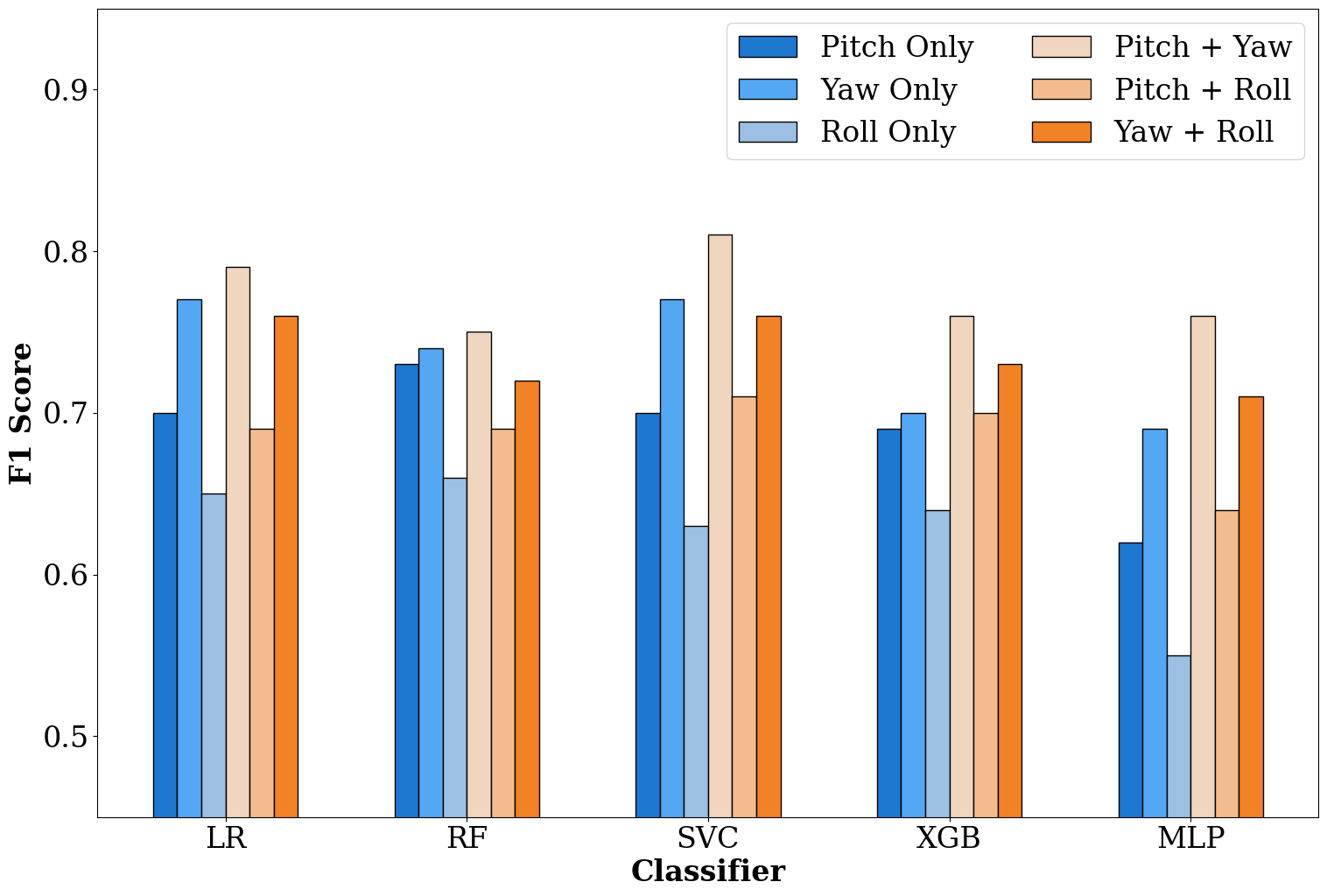}\hspace{0.15cm}\includegraphics[width=0.45\linewidth]{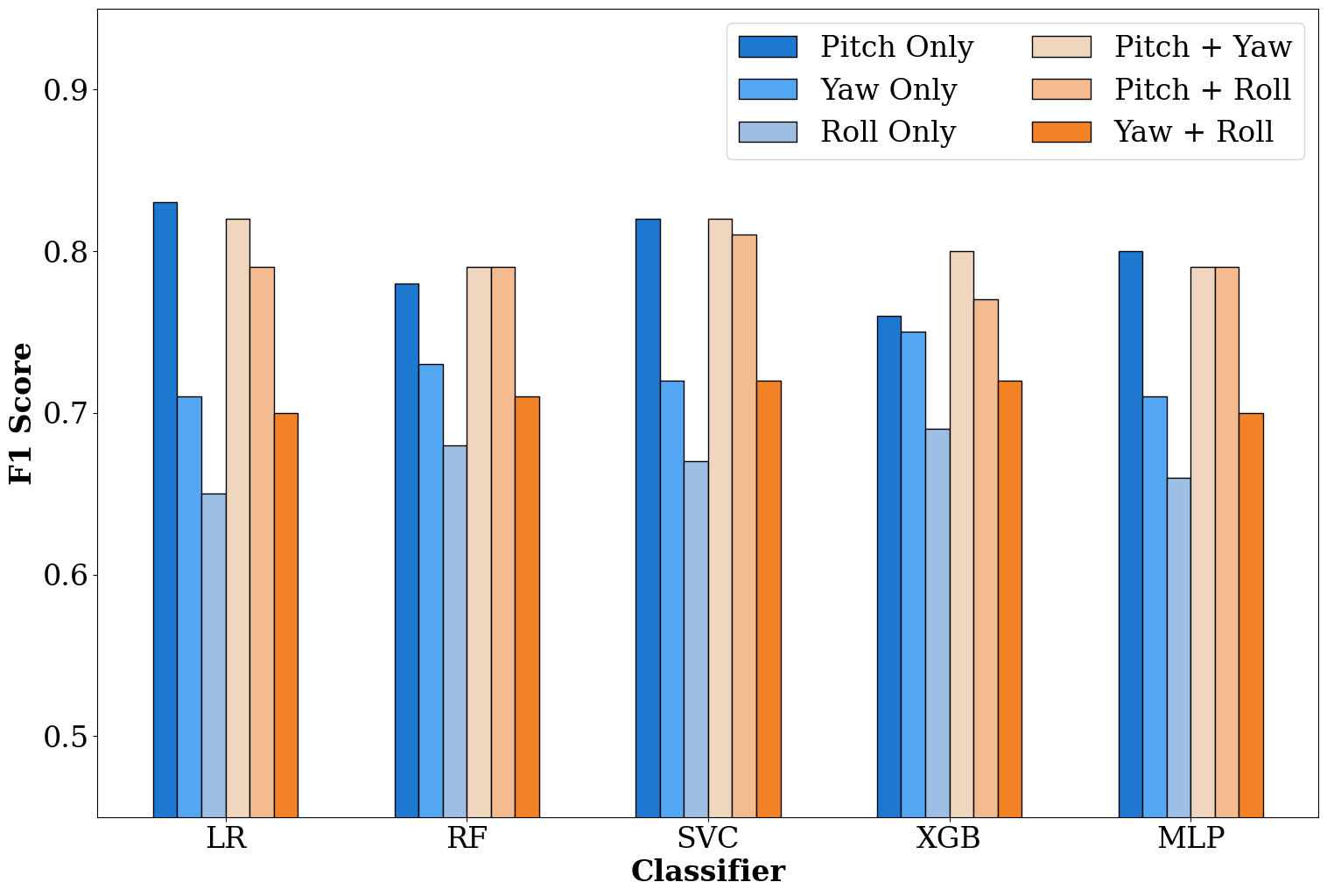}
        \vspace{-4mm}
\caption{Comparing F1 scores with different descriptors for the BlackDog (left) and AVEC2013 (right) datasets across classifiers.
  }\label{fig:descriptive_comp}\vspace{-4mm}
\end{figure*}

%
%
\subsection{\textbf{Ablative Analysis over Thin Slices}}
\label{sec:Ablative_ts}
Tables~\ref{tab:BDI_res1} and~\ref{tab:AVEC_res1} evaluate detection performance over (\emph{thin-slice}) chunks or short behavioural episodes, and over the entire video, on the BlackDog and AVEC2013 datasets. We further compared labelling performance at the chunk and video-levels using chunks spanning $15-135s$. The corresponding results are presented in Figure~\ref{fig:Chunk_vs_Vid_kineme}. For both plots presented in the figure, the dotted curves denote video-level F1-scores, while solid curves denote chunk-level scores obtained for different classifiers.

For the BlackDog dataset (Fig.~\ref{fig:Chunk_vs_Vid_kineme} (left)), longer time-slices (of length $75-105s$) achieve better performance than shorter ($15-60s$ long) ones at both the chunk and video-levels across all classifiers; these findings are consistent with the finding that more reliable predictions can be achieved with longer observations in general~\cite{madan_gahalawat_guha_subramanian_ICMI2021_Kinemes}. However, a performance drop is noted for very long chunk-lengths of $120-135s$ duration. Decoding results on the AVEC2013 dataset, consistent with Table~\ref{fig:kinemes_avec} results, a clear gap is noted between the chunk and video-level results, with the latter demonstrating superior performance. Very similar F1-scores are observed across classifiers for various chunk lengths. No clear trends are discernible from video-level F1-scores obtained with different chunk-lengths, except that the performance in general decreases for all classifiers with very long chunks.


%
%
\subsection{\textbf{Ablative Analysis over Angular Dimensions}}\label{sec:Ablative_ad}
To investigate the impact of the head pose angular dimensions on chunk-level binary depression detection performance, we perform detection utilising ($8 \times 1$) statistical features over each of the (pitch, yaw, and roll) angular dimensions, and concatenate features ($8 \times 2$) for the dimensional pairs to evaluate which angular dimension(s) are more informative.

Figure~\ref{fig:descriptive_comp} presents F1-scores obtained with the different classifiers for uni-dimensional and pairwise-dimensional features. On the BlackDog dataset, a combination of the pitch and yaw-based descriptors produce the best performance across all models, while roll-specific descriptors perform worst. For the AVEC2013 dataset, pitch-based descriptors achieve excellent performance across models. The F1-scores achieved with these features are very comparable to the pitch + yaw and pitch + roll combinations. Here again, roll-specific features achieve the worst performance. Cumulatively, these results convey that pitch is the most informative head pose dimension, with roll being the least informative. With respect to combinations, the pitch + yaw combination in general produces the best results. These results again confirm that responsiveness in social interactions, as captured by pitch (capturing actions such as head nodding) and yaw (capturing head shaking), provides a critical cue for detecting depression, consistent with prior studies ~\cite{hale1997non, alghowinem2013head}.

\section{Conclusion}\label{Sec:DC}

In this paper, we demonstrate the efficacy of elementary head motion units, termed \emph{kinemes}, for depression detection by utilising two approaches: (a) discovering kinemes from data of both patient and control cohorts, and (b) learning kineme patterns solely from the control cohort to compute statistical functional features derived from reconstruction errors for the two classes. Apart from effective depression detection, we also identify explainable kineme patterns for the two classes, consistent with prior research. 

Our study demonstrates the utility of head motion features for detecting depression, but our experiments are restricted to classification tasks involving a discretisation of the depression scores. In the future, we will investigate (a) the utility of kinemes for continuous prediction (regression) of depression severity, (b) the cross-dataset generalisability of models trained via kinemes, and (c) the development of multimodal methodologies combining kinemes with other behavioural markers, and evaluating their efficacy.


\bibliographystyle{ACM-Reference-Format}
\bibliography{references}


\begin{thebibliography}{45}


\ifx \showCODEN    \undefined \def \showCODEN     #1{\unskip}     \fi
\ifx \showDOI      \undefined \def \showDOI       #1{#1}\fi
\ifx \showISBNx    \undefined \def \showISBNx     #1{\unskip}     \fi
\ifx \showISBNxiii \undefined \def \showISBNxiii  #1{\unskip}     \fi
\ifx \showISSN     \undefined \def \showISSN      #1{\unskip}     \fi
\ifx \showLCCN     \undefined \def \showLCCN      #1{\unskip}     \fi
\ifx \shownote     \undefined \def \shownote      #1{#1}          \fi
\ifx \showarticletitle \undefined \def \showarticletitle #1{#1}   \fi
\ifx \showURL      \undefined \def \showURL       {\relax}        \fi
\providecommand\bibfield[2]{#2}
\providecommand\bibinfo[2]{#2}
\providecommand\natexlab[1]{#1}
\providecommand\showeprint[2][]{arXiv:#2}

\bibitem[\protect\citeauthoryear{Aguilera, Far{\'\i}as, Ortega-Mendoza, and
  Montes-y G{\'o}mez}{Aguilera et~al\mbox{.}}{2021}]%
        {aguilera2021depression}
\bibfield{author}{\bibinfo{person}{Juan Aguilera}, \bibinfo{person}{Delia
  Iraz{\'u}~Hern{\'a}ndez Far{\'\i}as}, \bibinfo{person}{Rosa~Mar{\'\i}a
  Ortega-Mendoza}, {and} \bibinfo{person}{Manuel Montes-y G{\'o}mez}.}
  \bibinfo{year}{2021}\natexlab{}.
\newblock \showarticletitle{Depression and anorexia detection in social media
  as a one-class classification problem}.
\newblock \bibinfo{journal}{\emph{Applied Intelligence}}  \bibinfo{volume}{51}
  (\bibinfo{year}{2021}), \bibinfo{pages}{6088--6103}.
\newblock


\bibitem[\protect\citeauthoryear{Al-gawwam and Benaissa}{Al-gawwam and
  Benaissa}{2018}]%
        {al2018depression}
\bibfield{author}{\bibinfo{person}{Sarmad Al-gawwam} {and}
  \bibinfo{person}{Mohammed Benaissa}.} \bibinfo{year}{2018}\natexlab{}.
\newblock \showarticletitle{Depression detection from eye blink features}. In
  \bibinfo{booktitle}{\emph{2018 IEEE international symposium on signal
  processing and information technology (ISSPIT)}}. IEEE,
  \bibinfo{pages}{388--392}.
\newblock


\bibitem[\protect\citeauthoryear{Alghowinem, Goecke, Cohn, Wagner, Parker, and
  Breakspear}{Alghowinem et~al\mbox{.}}{2015}]%
        {alghowinem2015cross}
\bibfield{author}{\bibinfo{person}{Sharifa Alghowinem}, \bibinfo{person}{Roland
  Goecke}, \bibinfo{person}{Jeffrey~F Cohn}, \bibinfo{person}{Michael Wagner},
  \bibinfo{person}{Gordon Parker}, {and} \bibinfo{person}{Michael Breakspear}.}
  \bibinfo{year}{2015}\natexlab{}.
\newblock \showarticletitle{Cross-cultural detection of depression from
  nonverbal behaviour}. In \bibinfo{booktitle}{\emph{2015 11th IEEE
  International conference and workshops on automatic face and gesture
  recognition (FG)}}, Vol.~\bibinfo{volume}{1}. IEEE, \bibinfo{pages}{1--8}.
\newblock


\bibitem[\protect\citeauthoryear{Alghowinem, Goecke, Wagner, Epps, Hyett,
  Parker, and Breakspear}{Alghowinem et~al\mbox{.}}{2016}]%
        {alghowinem2016multimodal}
\bibfield{author}{\bibinfo{person}{Sharifa Alghowinem}, \bibinfo{person}{Roland
  Goecke}, \bibinfo{person}{Michael Wagner}, \bibinfo{person}{Julien Epps},
  \bibinfo{person}{Matthew Hyett}, \bibinfo{person}{Gordon Parker}, {and}
  \bibinfo{person}{Michael Breakspear}.} \bibinfo{year}{2016}\natexlab{}.
\newblock \showarticletitle{{Multimodal depression detection: Fusion analysis
  of paralinguistic, head pose and eye gaze behaviors}}.
\newblock \bibinfo{journal}{\emph{IEEE Transactions on Affective Computing}}
  \bibinfo{volume}{9}, \bibinfo{number}{4} (\bibinfo{year}{2016}),
  \bibinfo{pages}{478--490}.
\newblock


\bibitem[\protect\citeauthoryear{Alghowinem, Goecke, Wagner, Parkerx, and
  Breakspear}{Alghowinem et~al\mbox{.}}{2013}]%
        {alghowinem2013head}
\bibfield{author}{\bibinfo{person}{Sharifa Alghowinem}, \bibinfo{person}{Roland
  Goecke}, \bibinfo{person}{Michael Wagner}, \bibinfo{person}{Gordon Parkerx},
  {and} \bibinfo{person}{Michael Breakspear}.} \bibinfo{year}{2013}\natexlab{}.
\newblock \showarticletitle{Head pose and movement analysis as an indicator of
  depression}. In \bibinfo{booktitle}{\emph{2013 Humaine Association Conference
  on Affective Computing and Intelligent Interaction}}. IEEE,
  \bibinfo{pages}{283--288}.
\newblock


\bibitem[\protect\citeauthoryear{Alghowinem, Gedeon, Goecke, Cohn, and
  Parker}{Alghowinem et~al\mbox{.}}{2020}]%
        {alghowinem2020interpretation}
\bibfield{author}{\bibinfo{person}{Sharifa~Mohammed Alghowinem},
  \bibinfo{person}{Tom Gedeon}, \bibinfo{person}{Roland Goecke},
  \bibinfo{person}{Jeffrey Cohn}, {and} \bibinfo{person}{Gordon Parker}.}
  \bibinfo{year}{2020}\natexlab{}.
\newblock \showarticletitle{Interpretation of depression detection models via
  feature selection methods}.
\newblock \bibinfo{journal}{\emph{IEEE transactions on affective computing}}
  (\bibinfo{year}{2020}).
\newblock


\bibitem[\protect\citeauthoryear{Baltrušaitis, Robinson, and
  Morency}{Baltrušaitis et~al\mbox{.}}{2016}]%
        {Baltrusaitis16}
\bibfield{author}{\bibinfo{person}{Tadas Baltrušaitis}, \bibinfo{person}{Peter
  Robinson}, {and} \bibinfo{person}{Louis-Philippe Morency}.}
  \bibinfo{year}{2016}\natexlab{}.
\newblock \showarticletitle{{OpenFace: An open source facial behavior analysis
  toolkit}}. In \bibinfo{booktitle}{\emph{2016 IEEE Winter Conference on
  Applications of Computer Vision (WACV)}}. \bibinfo{pages}{1--10}.
\newblock
\urldef\tempurl%
\url{https://doi.org/10.1109/WACV.2016.7477553}
\showDOI{\tempurl}


\bibitem[\protect\citeauthoryear{Beck, Steer, Ball, and Ranieri}{Beck
  et~al\mbox{.}}{1996}]%
        {beck1996comparison}
\bibfield{author}{\bibinfo{person}{Aaron~T Beck}, \bibinfo{person}{Robert~A
  Steer}, \bibinfo{person}{Roberta Ball}, {and} \bibinfo{person}{William~F
  Ranieri}.} \bibinfo{year}{1996}\natexlab{}.
\newblock \showarticletitle{Comparison of Beck Depression Inventories-IA and-II
  in psychiatric outpatients}.
\newblock \bibinfo{journal}{\emph{Journal of personality assessment}}
  \bibinfo{volume}{67}, \bibinfo{number}{3} (\bibinfo{year}{1996}),
  \bibinfo{pages}{588--597}.
\newblock


\bibitem[\protect\citeauthoryear{Bourke, Douglas, and Porter}{Bourke
  et~al\mbox{.}}{2010}]%
        {bourke2010processing}
\bibfield{author}{\bibinfo{person}{Cecilia Bourke}, \bibinfo{person}{Katie
  Douglas}, {and} \bibinfo{person}{Richard Porter}.}
  \bibinfo{year}{2010}\natexlab{}.
\newblock \showarticletitle{Processing of facial emotion expression in major
  depression: a review}.
\newblock \bibinfo{journal}{\emph{Australian \& New Zealand Journal of
  Psychiatry}} \bibinfo{volume}{44}, \bibinfo{number}{8}
  (\bibinfo{year}{2010}), \bibinfo{pages}{681--696}.
\newblock


\bibitem[\protect\citeauthoryear{Campayo, G{\'o}mez-Biel, and Lobo}{Campayo
  et~al\mbox{.}}{2011}]%
        {campayo2011diabetes}
\bibfield{author}{\bibinfo{person}{Antonio Campayo}, \bibinfo{person}{Carlos~H
  G{\'o}mez-Biel}, {and} \bibinfo{person}{Antonio Lobo}.}
  \bibinfo{year}{2011}\natexlab{}.
\newblock \showarticletitle{Diabetes and depression}.
\newblock \bibinfo{journal}{\emph{Current psychiatry reports}}
  \bibinfo{volume}{13}, \bibinfo{number}{1} (\bibinfo{year}{2011}),
  \bibinfo{pages}{26--30}.
\newblock


\bibitem[\protect\citeauthoryear{Cohn, Cummins, Epps, Goecke, Joshi, and
  Scherer}{Cohn et~al\mbox{.}}{2018}]%
        {cohn2018multimodal}
\bibfield{author}{\bibinfo{person}{Jeffrey~F Cohn}, \bibinfo{person}{Nicholas
  Cummins}, \bibinfo{person}{Julien Epps}, \bibinfo{person}{Roland Goecke},
  \bibinfo{person}{Jyoti Joshi}, {and} \bibinfo{person}{Stefan Scherer}.}
  \bibinfo{year}{2018}\natexlab{}.
\newblock \showarticletitle{Multimodal assessment of depression from behavioral
  signals}.
\newblock \bibinfo{journal}{\emph{The Handbook of Multimodal-Multisensor
  Interfaces: Signal Processing, Architectures, and Detection of Emotion and
  Cognition-Volume 2}} (\bibinfo{year}{2018}), \bibinfo{pages}{375--417}.
\newblock


\bibitem[\protect\citeauthoryear{Cummins, Epps, Breakspear, and Goecke}{Cummins
  et~al\mbox{.}}{2011}]%
        {cummins2011investigation}
\bibfield{author}{\bibinfo{person}{Nicholas Cummins}, \bibinfo{person}{Julien
  Epps}, \bibinfo{person}{Michael Breakspear}, {and} \bibinfo{person}{Roland
  Goecke}.} \bibinfo{year}{2011}\natexlab{}.
\newblock \showarticletitle{An investigation of depressed speech detection:
  Features and normalization}. In \bibinfo{booktitle}{\emph{Twelfth Annual
  Conference of the International Speech Communication Association}}.
\newblock


\bibitem[\protect\citeauthoryear{de~Melo, Granger, and Hadid}{de~Melo
  et~al\mbox{.}}{2019}]%
        {de2019combining}
\bibfield{author}{\bibinfo{person}{Wheidima~Carneiro de Melo},
  \bibinfo{person}{Eric Granger}, {and} \bibinfo{person}{Abdenour Hadid}.}
  \bibinfo{year}{2019}\natexlab{}.
\newblock \showarticletitle{Combining global and local convolutional 3d
  networks for detecting depression from facial expressions}. In
  \bibinfo{booktitle}{\emph{2019 14th ieee international conference on
  automatic face \& gesture recognition (fg 2019)}}. IEEE,
  \bibinfo{pages}{1--8}.
\newblock


\bibitem[\protect\citeauthoryear{Dibeklio{\u{g}}lu, Hammal, and
  Cohn}{Dibeklio{\u{g}}lu et~al\mbox{.}}{2017}]%
        {dibekliouglu2017dynamic}
\bibfield{author}{\bibinfo{person}{Hamdi Dibeklio{\u{g}}lu},
  \bibinfo{person}{Zakia Hammal}, {and} \bibinfo{person}{Jeffrey~F Cohn}.}
  \bibinfo{year}{2017}\natexlab{}.
\newblock \showarticletitle{Dynamic multimodal measurement of depression
  severity using deep autoencoding}.
\newblock \bibinfo{journal}{\emph{IEEE journal of biomedical and health
  informatics}} \bibinfo{volume}{22}, \bibinfo{number}{2}
  (\bibinfo{year}{2017}), \bibinfo{pages}{525--536}.
\newblock


\bibitem[\protect\citeauthoryear{Dibeklio{\u{g}}lu, Hammal, Yang, and
  Cohn}{Dibeklio{\u{g}}lu et~al\mbox{.}}{2015}]%
        {dibekliouglu2015multimodal}
\bibfield{author}{\bibinfo{person}{Hamdi Dibeklio{\u{g}}lu},
  \bibinfo{person}{Zakia Hammal}, \bibinfo{person}{Ying Yang}, {and}
  \bibinfo{person}{Jeffrey~F Cohn}.} \bibinfo{year}{2015}\natexlab{}.
\newblock \showarticletitle{Multimodal detection of depression in clinical
  interviews}. In \bibinfo{booktitle}{\emph{Proceedings of the 2015 ACM on
  international conference on multimodal interaction}}.
  \bibinfo{pages}{307--310}.
\newblock


\bibitem[\protect\citeauthoryear{Fossi, Faravelli, and Paoli}{Fossi
  et~al\mbox{.}}{1984}]%
        {fossi1984ethological}
\bibfield{author}{\bibinfo{person}{Luciano Fossi}, \bibinfo{person}{C
  Faravelli}, {and} \bibinfo{person}{M Paoli}.}
  \bibinfo{year}{1984}\natexlab{}.
\newblock \showarticletitle{The ethological approach to the assessment of
  depressive disorders}.
\newblock \bibinfo{journal}{\emph{The Journal of nervous and mental disease}}
  \bibinfo{volume}{172}, \bibinfo{number}{6} (\bibinfo{year}{1984}),
  \bibinfo{pages}{332--341}.
\newblock


\bibitem[\protect\citeauthoryear{Gerych, Agu, and Rundensteiner}{Gerych
  et~al\mbox{.}}{2019}]%
        {gerych2019classifying}
\bibfield{author}{\bibinfo{person}{Walter Gerych}, \bibinfo{person}{Emmanuel
  Agu}, {and} \bibinfo{person}{Elke Rundensteiner}.}
  \bibinfo{year}{2019}\natexlab{}.
\newblock \showarticletitle{Classifying depression in imbalanced datasets using
  an autoencoder-based anomaly detection approach}. In
  \bibinfo{booktitle}{\emph{2019 IEEE 13th International Conference on Semantic
  Computing (ICSC)}}. IEEE, \bibinfo{pages}{124--127}.
\newblock


\bibitem[\protect\citeauthoryear{Goldney, Wilson, Grande, Fisher, and
  McFarlane}{Goldney et~al\mbox{.}}{2000}]%
        {goldney2000suicidal}
\bibfield{author}{\bibinfo{person}{Robert~D Goldney}, \bibinfo{person}{David
  Wilson}, \bibinfo{person}{Eleonora~Dal Grande}, \bibinfo{person}{Laura~J
  Fisher}, {and} \bibinfo{person}{Alexander~C McFarlane}.}
  \bibinfo{year}{2000}\natexlab{}.
\newblock \showarticletitle{Suicidal ideation in a random community sample:
  attributable risk due to depression and psychosocial and traumatic events}.
\newblock \bibinfo{journal}{\emph{Australian \& New Zealand Journal of
  Psychiatry}} \bibinfo{volume}{34}, \bibinfo{number}{1}
  (\bibinfo{year}{2000}), \bibinfo{pages}{98--106}.
\newblock


\bibitem[\protect\citeauthoryear{Greenberg, Fournier, Sisitsky, Pike, and
  Kessler}{Greenberg et~al\mbox{.}}{2015}]%
        {greenberg2015economic}
\bibfield{author}{\bibinfo{person}{Paul~E Greenberg},
  \bibinfo{person}{Andree-Anne Fournier}, \bibinfo{person}{Tammy Sisitsky},
  \bibinfo{person}{Crystal~T Pike}, {and} \bibinfo{person}{Ronald~C Kessler}.}
  \bibinfo{year}{2015}\natexlab{}.
\newblock \showarticletitle{The economic burden of adults with major depressive
  disorder in the United States (2005 and 2010)}.
\newblock \bibinfo{journal}{\emph{The Journal of clinical psychiatry}}
  \bibinfo{volume}{76}, \bibinfo{number}{2} (\bibinfo{year}{2015}),
  \bibinfo{pages}{5356}.
\newblock


\bibitem[\protect\citeauthoryear{Hale~III, Jansen, Bouhuys, Jenner, and van~den
  Hoofdakker}{Hale~III et~al\mbox{.}}{1997}]%
        {hale1997non}
\bibfield{author}{\bibinfo{person}{William~W Hale~III},
  \bibinfo{person}{Jaap~HC Jansen}, \bibinfo{person}{Antoinette~L Bouhuys},
  \bibinfo{person}{Jack~A Jenner}, {and} \bibinfo{person}{Rutger~H van~den
  Hoofdakker}.} \bibinfo{year}{1997}\natexlab{}.
\newblock \showarticletitle{Non-verbal behavioral interactions of depressed
  patients with partners and strangers: The role of behavioral social support
  and involvement in depression persistence}.
\newblock \bibinfo{journal}{\emph{Journal of affective disorders}}
  \bibinfo{volume}{44}, \bibinfo{number}{2-3} (\bibinfo{year}{1997}),
  \bibinfo{pages}{111--122}.
\newblock


\bibitem[\protect\citeauthoryear{He, Guo, Tiwari, Pandey, and Dang}{He
  et~al\mbox{.}}{2022}]%
        {he2022intelligent}
\bibfield{author}{\bibinfo{person}{Lang He}, \bibinfo{person}{Chenguang Guo},
  \bibinfo{person}{Prayag Tiwari}, \bibinfo{person}{Hari~Mohan Pandey}, {and}
  \bibinfo{person}{Wei Dang}.} \bibinfo{year}{2022}\natexlab{}.
\newblock \showarticletitle{Intelligent system for depression scale estimation
  with facial expressions and case study in industrial intelligence}.
\newblock \bibinfo{journal}{\emph{International Journal of Intelligent
  Systems}} \bibinfo{volume}{37}, \bibinfo{number}{12} (\bibinfo{year}{2022}),
  \bibinfo{pages}{10140--10156}.
\newblock


\bibitem[\protect\citeauthoryear{Huang, Epps, and Joachim}{Huang
  et~al\mbox{.}}{2019}]%
        {huang2019investigation}
\bibfield{author}{\bibinfo{person}{Zhaocheng Huang}, \bibinfo{person}{Julien
  Epps}, {and} \bibinfo{person}{Dale Joachim}.}
  \bibinfo{year}{2019}\natexlab{}.
\newblock \showarticletitle{Investigation of speech landmark patterns for
  depression detection}.
\newblock \bibinfo{journal}{\emph{IEEE Transactions on Affective Computing}}
  \bibinfo{volume}{13}, \bibinfo{number}{2} (\bibinfo{year}{2019}),
  \bibinfo{pages}{666--679}.
\newblock


\bibitem[\protect\citeauthoryear{Joshi, Dhall, Goecke, and Cohn}{Joshi
  et~al\mbox{.}}{2013a}]%
        {joshi2013relative}
\bibfield{author}{\bibinfo{person}{Jyoti Joshi}, \bibinfo{person}{Abhinav
  Dhall}, \bibinfo{person}{Roland Goecke}, {and} \bibinfo{person}{Jeffrey~F
  Cohn}.} \bibinfo{year}{2013}\natexlab{a}.
\newblock \showarticletitle{Relative body parts movement for automatic
  depression analysis}. In \bibinfo{booktitle}{\emph{2013 Humaine association
  conference on affective computing and intelligent interaction}}. IEEE,
  \bibinfo{pages}{492--497}.
\newblock


\bibitem[\protect\citeauthoryear{Joshi, Goecke, Parker, and Breakspear}{Joshi
  et~al\mbox{.}}{2013b}]%
        {joshi2013can}
\bibfield{author}{\bibinfo{person}{Jyoti Joshi}, \bibinfo{person}{Roland
  Goecke}, \bibinfo{person}{Gordon Parker}, {and} \bibinfo{person}{Michael
  Breakspear}.} \bibinfo{year}{2013}\natexlab{b}.
\newblock \showarticletitle{Can body expressions contribute to automatic
  depression analysis?}. In \bibinfo{booktitle}{\emph{2013 10th IEEE
  International Conference and Workshops on Automatic Face and Gesture
  Recognition (FG)}}. IEEE, \bibinfo{pages}{1--7}.
\newblock


\bibitem[\protect\citeauthoryear{Kacem, Hammal, Daoudi, and Cohn}{Kacem
  et~al\mbox{.}}{2018}]%
        {kacem2018detecting}
\bibfield{author}{\bibinfo{person}{Anis Kacem}, \bibinfo{person}{Zakia Hammal},
  \bibinfo{person}{Mohamed Daoudi}, {and} \bibinfo{person}{Jeffrey Cohn}.}
  \bibinfo{year}{2018}\natexlab{}.
\newblock \showarticletitle{Detecting depression severity by interpretable
  representations of motion dynamics}. In \bibinfo{booktitle}{\emph{2018 13th
  ieee international conference on automatic face \& gesture recognition (fg
  2018)}}. IEEE, \bibinfo{pages}{739--745}.
\newblock


\bibitem[\protect\citeauthoryear{L{\'e}pine and Briley}{L{\'e}pine and
  Briley}{2011}]%
        {lepine2011increasing}
\bibfield{author}{\bibinfo{person}{Jean-Pierre L{\'e}pine} {and}
  \bibinfo{person}{Mike Briley}.} \bibinfo{year}{2011}\natexlab{}.
\newblock \showarticletitle{The increasing burden of depression}.
\newblock \bibinfo{journal}{\emph{Neuropsychiatric disease and treatment}}
  \bibinfo{volume}{7}, \bibinfo{number}{sup1} (\bibinfo{year}{2011}),
  \bibinfo{pages}{3--7}.
\newblock


\bibitem[\protect\citeauthoryear{Madan, Gahalawat, Guha, and Subramanian}{Madan
  et~al\mbox{.}}{2021}]%
        {madan_gahalawat_guha_subramanian_ICMI2021_Kinemes}
\bibfield{author}{\bibinfo{person}{Surbhi Madan}, \bibinfo{person}{Monika
  Gahalawat}, \bibinfo{person}{Tanaya Guha}, {and} \bibinfo{person}{Ramanathan
  Subramanian}.} \bibinfo{year}{2021}\natexlab{}.
\newblock \showarticletitle{{Head Matters: Explainable Human-Centered Trait
  Prediction from Head Motion Dynamics}}. In
  \bibinfo{booktitle}{\emph{Proceedings of the 2021 International Conference on
  Multimodal Interaction}} (Montr\'{e}al, QC, Canada)
  \emph{(\bibinfo{series}{ICMI '21})}. \bibinfo{publisher}{Association for
  Computing Machinery}, \bibinfo{address}{New York, NY, USA},
  \bibinfo{pages}{435–--443}.
\newblock
\urldef\tempurl%
\url{https://doi.org/10.1145/3462244.3479901}
\showDOI{\tempurl}


\bibitem[\protect\citeauthoryear{Morales, Scherer, and Levitan}{Morales
  et~al\mbox{.}}{2017}]%
        {morales2017cross}
\bibfield{author}{\bibinfo{person}{Michelle Morales}, \bibinfo{person}{Stefan
  Scherer}, {and} \bibinfo{person}{Rivka Levitan}.}
  \bibinfo{year}{2017}\natexlab{}.
\newblock \showarticletitle{A cross-modal review of indicators for depression
  detection systems}. In \bibinfo{booktitle}{\emph{Proceedings of the fourth
  workshop on computational linguistics and clinical psychology—From
  linguistic signal to clinical reality}}. \bibinfo{pages}{1--12}.
\newblock


\bibitem[\protect\citeauthoryear{Mour{\~a}o-Miranda, Hardoon, Hahn, Marquand,
  Williams, Shawe-Taylor, and Brammer}{Mour{\~a}o-Miranda
  et~al\mbox{.}}{2011}]%
        {mourao2011patient}
\bibfield{author}{\bibinfo{person}{Janaina Mour{\~a}o-Miranda},
  \bibinfo{person}{David~R Hardoon}, \bibinfo{person}{Tim Hahn},
  \bibinfo{person}{Andre~F Marquand}, \bibinfo{person}{Steve~CR Williams},
  \bibinfo{person}{John Shawe-Taylor}, {and} \bibinfo{person}{Michael
  Brammer}.} \bibinfo{year}{2011}\natexlab{}.
\newblock \showarticletitle{Patient classification as an outlier detection
  problem: an application of the one-class support vector machine}.
\newblock \bibinfo{journal}{\emph{Neuroimage}} \bibinfo{volume}{58},
  \bibinfo{number}{3} (\bibinfo{year}{2011}), \bibinfo{pages}{793--804}.
\newblock


\bibitem[\protect\citeauthoryear{Nasir, Jati, Shivakumar,
  Nallan~Chakravarthula, and Georgiou}{Nasir et~al\mbox{.}}{2016}]%
        {nasir2016multimodal}
\bibfield{author}{\bibinfo{person}{Md Nasir}, \bibinfo{person}{Arindam Jati},
  \bibinfo{person}{Prashanth~Gurunath Shivakumar}, \bibinfo{person}{Sandeep
  Nallan~Chakravarthula}, {and} \bibinfo{person}{Panayiotis Georgiou}.}
  \bibinfo{year}{2016}\natexlab{}.
\newblock \showarticletitle{Multimodal and multiresolution depression detection
  from speech and facial landmark features}. In
  \bibinfo{booktitle}{\emph{Proceedings of the 6th international workshop on
  audio/visual emotion challenge}}. \bibinfo{pages}{43--50}.
\newblock


\bibitem[\protect\citeauthoryear{of~Health~Metrics and
  Evaluation}{of~Health~Metrics and Evaluation}{2021}]%
        {institute2021global}
\bibfield{author}{\bibinfo{person}{Institute of Health~Metrics} {and}
  \bibinfo{person}{Evaluation}.} \bibinfo{year}{2021}\natexlab{}.
\newblock \bibinfo{title}{Global Health Data Exchange (GHDx)}.
\newblock
\newblock


\bibitem[\protect\citeauthoryear{Opoku~Asare, Visuri, and Ferreira}{Opoku~Asare
  et~al\mbox{.}}{2019}]%
        {opoku2019towards}
\bibfield{author}{\bibinfo{person}{Kennedy Opoku~Asare}, \bibinfo{person}{Aku
  Visuri}, {and} \bibinfo{person}{Denzil~ST Ferreira}.}
  \bibinfo{year}{2019}\natexlab{}.
\newblock \showarticletitle{Towards early detection of depression through
  smartphone sensing}. In \bibinfo{booktitle}{\emph{Adjunct Proceedings of the
  2019 ACM International Joint Conference on Pervasive and Ubiquitous Computing
  and Proceedings of the 2019 ACM International Symposium on Wearable
  Computers}}. \bibinfo{pages}{1158--1161}.
\newblock


\bibitem[\protect\citeauthoryear{Pampouchidou, Simos, Marias, Meriaudeau, Yang,
  Pediaditis, and Tsiknakis}{Pampouchidou et~al\mbox{.}}{2019}]%
        {pampouchidou_et_al_TAC_DepressionReview}
\bibfield{author}{\bibinfo{person}{Anastasia Pampouchidou},
  \bibinfo{person}{Panagiotis~G. Simos}, \bibinfo{person}{Kostas Marias},
  \bibinfo{person}{Fabrice Meriaudeau}, \bibinfo{person}{Fan Yang},
  \bibinfo{person}{Matthew Pediaditis}, {and} \bibinfo{person}{Manolis
  Tsiknakis}.} \bibinfo{year}{2019}\natexlab{}.
\newblock \showarticletitle{{Automatic Assessment of Depression Based on Visual
  Cues: A Systematic Review}}.
\newblock \bibinfo{journal}{\emph{IEEE Transactions on Affective Computing}}
  \bibinfo{volume}{10}, \bibinfo{number}{4} (\bibinfo{year}{2019}),
  \bibinfo{pages}{445--470}.
\newblock
\urldef\tempurl%
\url{https://doi.org/10.1109/TAFFC.2017.2724035}
\showDOI{\tempurl}


\bibitem[\protect\citeauthoryear{Parekh, Foong, Zhao, and Subramanian}{Parekh
  et~al\mbox{.}}{2018}]%
        {Parekh2018}
\bibfield{author}{\bibinfo{person}{Viral Parekh}, \bibinfo{person}{Pin~Sym
  Foong}, \bibinfo{person}{Shengdong Zhao}, {and} \bibinfo{person}{Ramanathan
  Subramanian}.} \bibinfo{year}{2018}\natexlab{}.
\newblock \showarticletitle{AVEID: Automatic Video System for Measuring
  Engagement In Dementia}. In \bibinfo{booktitle}{\emph{23rd International
  Conference on Intelligent User Interfaces}} (Tokyo, Japan)
  \emph{(\bibinfo{series}{IUI '18})}. \bibinfo{pages}{409–413}.
\newblock
\urldef\tempurl%
\url{https://doi.org/10.1145/3172944.3173010}
\showDOI{\tempurl}


\bibitem[\protect\citeauthoryear{Pedersen, Schelde, Hannibal, Behnke, Nielsen,
  and Hertz}{Pedersen et~al\mbox{.}}{1988}]%
        {pedersen1988ethological}
\bibfield{author}{\bibinfo{person}{Jesper Pedersen}, \bibinfo{person}{JTM
  Schelde}, \bibinfo{person}{E Hannibal}, \bibinfo{person}{K Behnke},
  \bibinfo{person}{BM Nielsen}, {and} \bibinfo{person}{M Hertz}.}
  \bibinfo{year}{1988}\natexlab{}.
\newblock \showarticletitle{An ethological description of depression}.
\newblock \bibinfo{journal}{\emph{Acta psychiatrica scandinavica}}
  \bibinfo{volume}{78}, \bibinfo{number}{3} (\bibinfo{year}{1988}),
  \bibinfo{pages}{320--330}.
\newblock


\bibitem[\protect\citeauthoryear{Rejaibi, Komaty, Meriaudeau, Agrebi, and
  Othmani}{Rejaibi et~al\mbox{.}}{2022}]%
        {rejaibi2022mfcc}
\bibfield{author}{\bibinfo{person}{Emna Rejaibi}, \bibinfo{person}{Ali Komaty},
  \bibinfo{person}{Fabrice Meriaudeau}, \bibinfo{person}{Said Agrebi}, {and}
  \bibinfo{person}{Alice Othmani}.} \bibinfo{year}{2022}\natexlab{}.
\newblock \showarticletitle{MFCC-based recurrent neural network for automatic
  clinical depression recognition and assessment from speech}.
\newblock \bibinfo{journal}{\emph{Biomedical Signal Processing and Control}}
  \bibinfo{volume}{71} (\bibinfo{year}{2022}), \bibinfo{pages}{103107}.
\newblock


\bibitem[\protect\citeauthoryear{Ringeval, Schuller, Valstar, Cummins, Cowie,
  Tavabi, Schmitt, Alisamir, Amiriparian, Messner, et~al\mbox{.}}{Ringeval
  et~al\mbox{.}}{2019}]%
        {ringeval2019avec}
\bibfield{author}{\bibinfo{person}{Fabien Ringeval}, \bibinfo{person}{Bj{\"o}rn
  Schuller}, \bibinfo{person}{Michel Valstar}, \bibinfo{person}{Nicholas
  Cummins}, \bibinfo{person}{Roddy Cowie}, \bibinfo{person}{Leili Tavabi},
  \bibinfo{person}{Maximilian Schmitt}, \bibinfo{person}{Sina Alisamir},
  \bibinfo{person}{Shahin Amiriparian}, \bibinfo{person}{Eva-Maria Messner},
  {et~al\mbox{.}}} \bibinfo{year}{2019}\natexlab{}.
\newblock \showarticletitle{AVEC 2019 workshop and challenge: state-of-mind,
  detecting depression with AI, and cross-cultural affect recognition}. In
  \bibinfo{booktitle}{\emph{Proceedings of the 9th International on
  Audio/visual Emotion Challenge and Workshop}}. \bibinfo{pages}{3--12}.
\newblock


\bibitem[\protect\citeauthoryear{Samanta and Guha}{Samanta and Guha}{2017}]%
        {samanta2017role}
\bibfield{author}{\bibinfo{person}{Atanu Samanta} {and} \bibinfo{person}{Tanaya
  Guha}.} \bibinfo{year}{2017}\natexlab{}.
\newblock \showarticletitle{On the role of head motion in affective
  expression}. In \bibinfo{booktitle}{\emph{IEEE International Conference on
  Acoustics, Speech and Signal Processing (ICASSP)}}. IEEE,
  \bibinfo{pages}{2886--2890}.
\newblock


\bibitem[\protect\citeauthoryear{Senoussaoui, Sarria-Paja, Santos, and
  Falk}{Senoussaoui et~al\mbox{.}}{2014}]%
        {senoussaoui2014model}
\bibfield{author}{\bibinfo{person}{Mohammed Senoussaoui},
  \bibinfo{person}{Milton Sarria-Paja}, \bibinfo{person}{Jo{\~a}o~F Santos},
  {and} \bibinfo{person}{Tiago~H Falk}.} \bibinfo{year}{2014}\natexlab{}.
\newblock \showarticletitle{Model fusion for multimodal depression
  classification and level detection}. In \bibinfo{booktitle}{\emph{Proceedings
  of the 4th International Workshop on Audio/Visual Emotion Challenge}}.
  \bibinfo{pages}{57--63}.
\newblock


\bibitem[\protect\citeauthoryear{Song, Jaiswal, Shen, and Valstar}{Song
  et~al\mbox{.}}{2020}]%
        {song2020spectral}
\bibfield{author}{\bibinfo{person}{Siyang Song}, \bibinfo{person}{Shashank
  Jaiswal}, \bibinfo{person}{Linlin Shen}, {and} \bibinfo{person}{Michel
  Valstar}.} \bibinfo{year}{2020}\natexlab{}.
\newblock \showarticletitle{Spectral representation of behaviour primitives for
  depression analysis}.
\newblock \bibinfo{journal}{\emph{IEEE Transactions on Affective Computing}}
  \bibinfo{volume}{13}, \bibinfo{number}{2} (\bibinfo{year}{2020}),
  \bibinfo{pages}{829--844}.
\newblock


\bibitem[\protect\citeauthoryear{Steffen, N{\"u}bel, Jacob, B{\"a}tzing, and
  Holstiege}{Steffen et~al\mbox{.}}{2020}]%
        {steffen_et_al_2020_BMCPsychiatry}
\bibfield{author}{\bibinfo{person}{Annika Steffen}, \bibinfo{person}{Julia
  N{\"u}bel}, \bibinfo{person}{Frank Jacob}, \bibinfo{person}{J{\"o}rg
  B{\"a}tzing}, {and} \bibinfo{person}{Jakob Holstiege}.}
  \bibinfo{year}{2020}\natexlab{}.
\newblock \showarticletitle{{Mental and somatic comorbidity of depression: a
  comprehensive cross-sectional analysis of 202 diagnosis groups using German
  nationwide ambulatory claims data}}.
\newblock \bibinfo{journal}{\emph{BMC Psychiatry}} (\bibinfo{date}{March}
  \bibinfo{year}{2020}).
\newblock


\bibitem[\protect\citeauthoryear{Valstar, Schuller, Smith, Almaev, Eyben,
  Krajewski, Cowie, and Pantic}{Valstar et~al\mbox{.}}{2014}]%
        {valstar2014avec}
\bibfield{author}{\bibinfo{person}{Michel Valstar}, \bibinfo{person}{Bj{\"o}rn
  Schuller}, \bibinfo{person}{Kirsty Smith}, \bibinfo{person}{Timur Almaev},
  \bibinfo{person}{Florian Eyben}, \bibinfo{person}{Jarek Krajewski},
  \bibinfo{person}{Roddy Cowie}, {and} \bibinfo{person}{Maja Pantic}.}
  \bibinfo{year}{2014}\natexlab{}.
\newblock \showarticletitle{Avec 2014: 3d dimensional affect and depression
  recognition challenge}. In \bibinfo{booktitle}{\emph{Proceedings of the 4th
  international workshop on audio/visual emotion challenge}}.
  \bibinfo{pages}{3--10}.
\newblock


\bibitem[\protect\citeauthoryear{Valstar, Schuller, Smith, Eyben, Jiang,
  Bilakhia, Schnieder, Cowie, and Pantic}{Valstar et~al\mbox{.}}{2013}]%
        {valstar2013avec}
\bibfield{author}{\bibinfo{person}{Michel Valstar}, \bibinfo{person}{Bj{\"o}rn
  Schuller}, \bibinfo{person}{Kirsty Smith}, \bibinfo{person}{Florian Eyben},
  \bibinfo{person}{Bihan Jiang}, \bibinfo{person}{Sanjay Bilakhia},
  \bibinfo{person}{Sebastian Schnieder}, \bibinfo{person}{Roddy Cowie}, {and}
  \bibinfo{person}{Maja Pantic}.} \bibinfo{year}{2013}\natexlab{}.
\newblock \showarticletitle{Avec 2013: the continuous audio/visual emotion and
  depression recognition challenge}. In \bibinfo{booktitle}{\emph{Proceedings
  of the 3rd ACM international workshop on Audio/visual emotion challenge}}.
  \bibinfo{pages}{3--10}.
\newblock


\bibitem[\protect\citeauthoryear{Vos, Allen, and Arora}{Vos
  et~al\mbox{.}}{2016}]%
        {vos_et_al_GBD2016}
\bibfield{author}{\bibinfo{person}{Theo Vos}, \bibinfo{person}{Christine
  Allen}, {and} \bibinfo{person}{Megha Arora}.}
  \bibinfo{year}{2016}\natexlab{}.
\newblock \showarticletitle{{Global, regional, and national incidence,
  prevalence, and years lived with disability for 310 diseases and injuries,
  1990–2015: a systematic analysis for the Global Burden of Disease Study
  2015}}.
\newblock \bibinfo{journal}{\emph{The Lancet}} \bibinfo{volume}{388},
  \bibinfo{number}{10053} (\bibinfo{year}{2016}), \bibinfo{pages}{1545--1602}.
\newblock
\urldef\tempurl%
\url{https://doi.org/10.1016/S0140-6736(16)31678-6}
\showDOI{\tempurl}


\bibitem[\protect\citeauthoryear{Waxer}{Waxer}{1974}]%
        {waxer1974nonverbal}
\bibfield{author}{\bibinfo{person}{Peter Waxer}.}
  \bibinfo{year}{1974}\natexlab{}.
\newblock \showarticletitle{Nonverbal cues for depression.}
\newblock \bibinfo{journal}{\emph{Journal of Abnormal Psychology}}
  \bibinfo{volume}{83}, \bibinfo{number}{3} (\bibinfo{year}{1974}),
  \bibinfo{pages}{319}.
\newblock


\end{thebibliography}

\end{document}